\documentclass[journal,final]{IEEEtran}

\usepackage{normalpkgs}
\usepackage[switch,columnwise]{lineno}
\usepackage{hyperref}
\usepackage{url}
\usepackage{xcolor}
\usepackage{multirow}
\usepackage{float}
\usepackage{graphicx}
\usepackage{subcaption}
\usepackage{threeparttable}
\usepackage{float}
\usepackage{framed,multirow}
\usepackage{pifont}
\usepackage{amsmath}
\modulolinenumbers[5]
\usepackage[compress]{cite}
\usepackage{soul}

\begin{document}
%

\title{Deep Adversarial Transition Learning using Cross-Grafted Generative Stacks}

\author{Jinyong Hou~\IEEEmembership{Student Member,~IEEE},
        Xuejie Ding,
        Stephen Cranefield,
        Jeremiah D. Deng~\IEEEmembership{Member,~IEEE}
\thanks{This work was was partially supported by the UORG Grant 2020 (Corresponding author: Jeremiah D. Deng).\protect\\
J. Hou, X. Ding, S. Cranefield, and J. D. Deng are with the Department of Information Science, University of Otago, Dunedin 9054, New Zealand (e-mail: jeremiah.deng@otago.ac.nz).} 

}

\markboth{}%
{Shell \MakeLowercase{\textit{et al.}}: Bare Demo of IEEEtran.cls for IEEE Journals}

\maketitle


\begin{abstract}
Current deep domain adaptation methods used in computer vision have mainly focused on learning discriminative and domain-invariant features across different domains. In this paper, we present a novel ``deep adversarial transition learning'' (DATL) framework that bridges the domain gap by projecting the source and target domains into intermediate, transitional spaces through the employment of adjustable, cross-grafted generative network stacks and effective adversarial learning between transitions. Specifically, we construct variational auto-encoders (VAE) for the two domains, and form bidirectional transitions by cross-grafting the VAEs' decoder stacks. Furthermore, generative adversarial networks (GAN) are employed for domain adaptation, mapping the target domain data to the known label space of the source domain. The overall adaptation process hence consists of three phases: feature representation learning by VAEs, transitions generation, and transitions alignment by GANs. Experimental results demonstrate that our method outperforms the state-of-the-art on a number of unsupervised domain adaptation benchmarks.
\end{abstract}

\begin{IEEEkeywords}
  Domain Adaptation, Variational Auto-Encoders, Generative Adversarial Networks, Transfer learning, Transitional spaces
\end{IEEEkeywords}


%
\IEEEpeerreviewmaketitle


\section{Introduction} \label{sec:introduction}

\IEEEPARstart {I}{n} machine learning, domain adaptation aims to transfer knowledge learned previously from one or more ``source'' tasks to a new but related ``target'' domain. As a special form of transfer learning, it aims to overcome the lack of labelled data commonly existing in computer vision tasks, by learning from the labelled data of the source domain and adapting the gained knowledge to automatically annotate unlabelled data in the target domain~\cite{Pan2010}. It may also be used to recognize unfamiliar objects in a dynamically changing environment in robotics~\cite{Hoffman2016a,Tzeng2020}. Therefore, in recent years, domain adaptation, especially unsupervised domain adaptation, has attracted increasing research interests~\cite{ Ben-David2010,Adel2015,Long2014,Gopalan2011,Yosinski2014,Sener2016,Herath2017}. 
In general, for domain adaptation to occur, it is assumed that the source and target domains are located in the same label space, but there is a domain bias. The challenge is to  find an effective mechanism to overcome the domain bias and successfully map the unlabeled targets to the label space. 

One way to achieve the knowledge transfer between domains is to construct intermediate domain representations that classification models may benefit from in addition to the source domain data for learning. When the gap between the source and target domains is significant, progressive generation and alignment of the intermediate domains will become necessary. To this end, we extend the self-associative mechanism in autoencoder models~\cite{Kramer92,Kingma2014} and attempt to utilize deep representations trained across domains so as to achieve better transfer learning. 
By employing a network that contains autoencoders constructed from different domains with shared or exchanged stacks and is further trained using adversarial learning for label alignment, it is hopeful that we can exploit these cross-domain representations to leverage the network's domain adaptation ability.

{To implement these ideas, we propose in this paper 
a novel Deep Adversarial Transition  Learning (DATL) framework that 
recruits deep unsupervised generative representations from both the source and target domains to construct intermediate representations, so called \textit{transitions}, which are further aligned by adversarial learning to enable robust training and testing occurring within the transition spaces. } Our approach is partly inspired by UNIT~\cite{Liu2017}, but also differs from the latter in two aspects: First, the generative layers in DATL are grafted rather than just shared as in UNIT; Second, adversarial learning is carried out not within the original domains, but in the transition spaces as ``bridges''. Consequently, training and testing of the DATL framework are more capable of handling adaptation tasks with significant domain gaps. 
Specifically, we first construct two parallel variational auto-encoders (VAE)~\cite{Kingma2014} to extract latent encodings of the source and target domains using shared encoder stacks. Then we generate bidirectional transitions by employing a novel mechanism of cross-grafting the generative stacks in the decoders, denoted by CGGS. Furthermore, a couple of generative adversarial networks (GAN)~\cite{Goodfellow2014} are employed to carry out label alignment between the transitions generated by inputs from the source and target domains, in order to achieve accurate classification outcome. 

Due to these treatments, we believe our proposed DATL framework gives a promising new method for unsupervised domain adaptation. Through building transition spaces, feature learning is carried out across domains, thus potentially reducing domain-dependence and increasing domain-invariance, while the adversarial networks may further push feature representations away from domain differences, contributing to robust domain adaptation performance. Also, the cross-grafting process on the generative stacks is symmetric to both domains, leading to consistent performance regardless of the adaptation direction, as revealed by our experiment results. Another advantage of DATL as demonstrated in our experiments is that the learned network components are rather transferable across different tasks, i.e. the framework pretrained for one task can be employed to a new task without much re-training, which is an attractive trait for real-world applications. 



This paper is extended from a conference publication~\cite{Hou2019a}, where the CGGS-based framework was first presented and some preliminary experiment results were provided.
Here in this work we give a detailed presentation of the entire framework, with the following additional technical and experimental contents introduced: (1) A theoretical explanation of the generation of transition spaces as obtained from a probabilistic weights perturbation perspective; (2) new experimental results on more benchmark datasets to expand the empirical evaluation of DATL against the state of the art, plus the new results on cross-task generalization; and (3) further analysis and visual evaluation of the DATL framework.      

The rest of the paper is organized as follows. Section~\ref{sec:related_work} gives a brief review of related work. In Section~\ref{sec:model}, we outline the overall structure of our proposed framework, introduce the generation of cross-domain transitions using CGGS, and present the learning metrics used by the DATL model. The experimental results are presented and discussed in Section~\ref{sec:experiments}. Finally, we conclude the paper in Section~\ref{sec:conslusion} with a discussion on future work.  

\section{Related Work} \label{sec:related_work}


Initially, traditional machine learning and statistical methods were used to approximate the distribution of the target to the source.  
A category of this is to use domain distance metrics, such as maximum mean discrepancy (MMD) to approximate the domains in Reproduce Hilbert Kernel Space (RHKS). In~\cite{Pan2011}, it is proposed to learn transfer components across domains in a RKHS using MMD. Transfer joint matching~\cite{Long2014} combines feature matching in MMD and instance re-weighting for domain adaptation. Reference~\cite{Adel2015} uses the probabilistic approach to define a classifier for the source, and then the similarity between the source and the target is utilized for the adaptation. CORAL~\cite{Sun2016} uses correlation alignment for domain adaptation. It first calculates the covariance matrices of the source data and target data, then whitens the source, and finally, recolors the whitened source using the target's covariance.

Some works of shallow methods use intermediate representations to transfer the knowledge learned from a previous task for the learning of the target task. Self-taught learning~\cite{Raina2007} first constructs a sparse coding space using unsupervised learning on a large volume of natural images, then the targets are projected into the space to improve the recognition performance. For the geodesic flow kernel~\cite{Gong2012,Gopalan2014}, the source and target datasets are embedded into a Grassman manifold, and then a geodesic flow is constructed between the two domains. As as result, a number of feature subspaces are sampled along the geodesic flow, and a kernel can be defined on the incremental feature space, allowing a domain-adapted classifier to be built for the target dataset. In~\cite{Zhang2019e}, the manifold criterion (local information) is combined with MMD (global regularizer) for the intermediate generations to guide transfer learning.     

Recent works have shown that domain adaptation by the means of deep neural networks (DNNs) have achieved impressive performance due to their strong feature-learning capacity. This provides a considerable improvement for cross-domain recognition tasks~\cite{Venkateswara2017,Long2015,Tzeng2015,Long2016b,Rozantsev2019,Liu2017,Bousmalis2016,Ghifary2016}. Intermediate domains are deployed by the deep models to carry out the adaption. DLID~\cite{Chopra2013} uses deep sparse learning to extract an interpolated representation from a set of intermediate datasets constructed by combining the source and target datasets using progressively varying proportions; the features from these intermediate datasets are then concatenated to train a classifier for domain adaption. One very recent work~\cite{Pizzati2019} collects web-crawled domain-related images as a bridge for image-to-image translation and domain adaptation. The new images are added into the original datasets to improve  performance.    

In the context of deep models, to make the invariant features of different domains to be close, one strategy is to utilize domain distance metrics, as the shallow methods do. For example, in references~\cite{Pan2011, Long2014}, MMD are used to approximate the target to the source domain in RHKS. Another approach is to adopt adversarial learning~\cite{Tzeng2017,Tzeng2014,Ganin2016,Liu2017,Bousmalis2017,Liu2016,Liu2018}. DANN~\cite{Ganin2016} employs a gradient reversal layer between the feature layer and the domain discriminator, causing feature representation to anti-learn the domain difference and hence adapt well to the target domain. ADDA~\cite{Tzeng2017} first trains a convolutional neural network (CNN) using the source dataset. An adversarial phase then follows, with the CNN assigned to the target for domain discriminator training, and the new target encoder CNN is finally combined with source classifier to achieve the adaptation. 

The models mentioned above focused on feature-level adaption. Recently, pixel-level adaptation has received attention due to the power of generative adversarial networks (GAN). The PixelDA framework~\cite{Bousmalis2017} generates synthetic images from source-domain images that are mapped to the target domain. A task classifier then is trained from the source and synthetic images using the source labels. The model GtA~\cite{Sankaranarayanan2018} utilizes the extracted features of domains for adaptation. The feature encoder was updated by an adversarial generative mechanism  AC-GAN~\cite{Odena2017}. The extracted features are concatenated with their labels (a fake label for target) as the input of AC-GAN. An additional classifier is trained using source features to test the adaptation performance by the target features. CyCADA~\cite{Hoffman2018} integrates the CycleGAN~\cite{Zhu2017} with consistency objectives to achieve feature- and pixel-level adaptation. The backbone of the model is CycleGAN, and then to keep the semantic information, a reconstruction loss function is added between the original and adapted images. Due to the additional objectives, it has a complex overall loss function to train the model. 

Pixel-level adaptation is also often connected with image-to-image translation. UNIT~\cite{Liu2017} introduces an unsupervised image-to-image translation framework based on a couple of variational auto-encoders (VAEs) and GANs. To achieve translation, a pair of corresponding images in different domains are mapped to a shared latent representation space. In~\cite{Murez2018}, a general image translation strategy is used for domain adaption. Adversarial learning is applied to align the latent encodings and the translated images in a cross-domain scenario. A cycle consistency objective is added to achieve better performance. In DAI2I~\cite{Chen2020}, an adaptive mapping procedure is used to translate the out-of-domain images by the image-to-image tranlation model trained by the source images.

Our proposed DATL framework combines two ideas from prior work: constructing intermediate cross-domain pixel-level representations, and employing adversarial networks for label alignment. It is different from the existing methods of intermediate domains, which only act unidirectionally. Specifically, DATL incorporates VAEs to learn feature representations, a cross-grafting step to generate bidirectional cross-domain transitions, and a generative adversarial approach that carries out the alignment and classification on the source-target transitions instead of the original domains. A detailed description of our framework is given in Section~\ref{sec:model}.


\section{Proposed Model} \label{sec:model}

Let us consider two domains: one is a source domain $\mathcal{D}_s$, with $n_s$ images $\mathbf{X}_s=\{\pmb{x}_{i}^{s}\}_{i=1}^{n_s}$ and their corresponding labels $\pmb{y}_s=\{y_i^s\}_{i=1}^{n_s}$; the other is a target domain $\mathcal{D}_t=\{\mathbf{X}_t, \pmb{y}_t\}$, where $n_t$ images $\mathbf{X}_t=\{\pmb{x}_{i}^{t}\}_{i=1}^{n_t}$ are available, but their labels $\pmb{y}_t=\{y_i^t\}_{i=1}^{n_t}$ are not. The source and target domains are drawn from joint distributions $\mathcal{P}(\mathbf{X}_s, \pmb{y}_s)$ and $\mathcal{Q}(\mathbf{X}_t, \pmb{y}_t)$, with a domain bias: $\mathcal{P} \neq \mathcal{Q}$. Our goal is to learn transition spaces bearing similarity to both domains, i.e.~some joint distribution between $\mathcal{P}$ and $\mathcal{Q}$ as a bridge for knowledge transfer, based on which the target images can be successfully classified. 

\subsection{Motivation}
\begin{figure}[thbp]
    \centering
    \begin{subfigure}[t]{0.23\textwidth}
        \centering
        \includegraphics[width=0.9\textwidth]{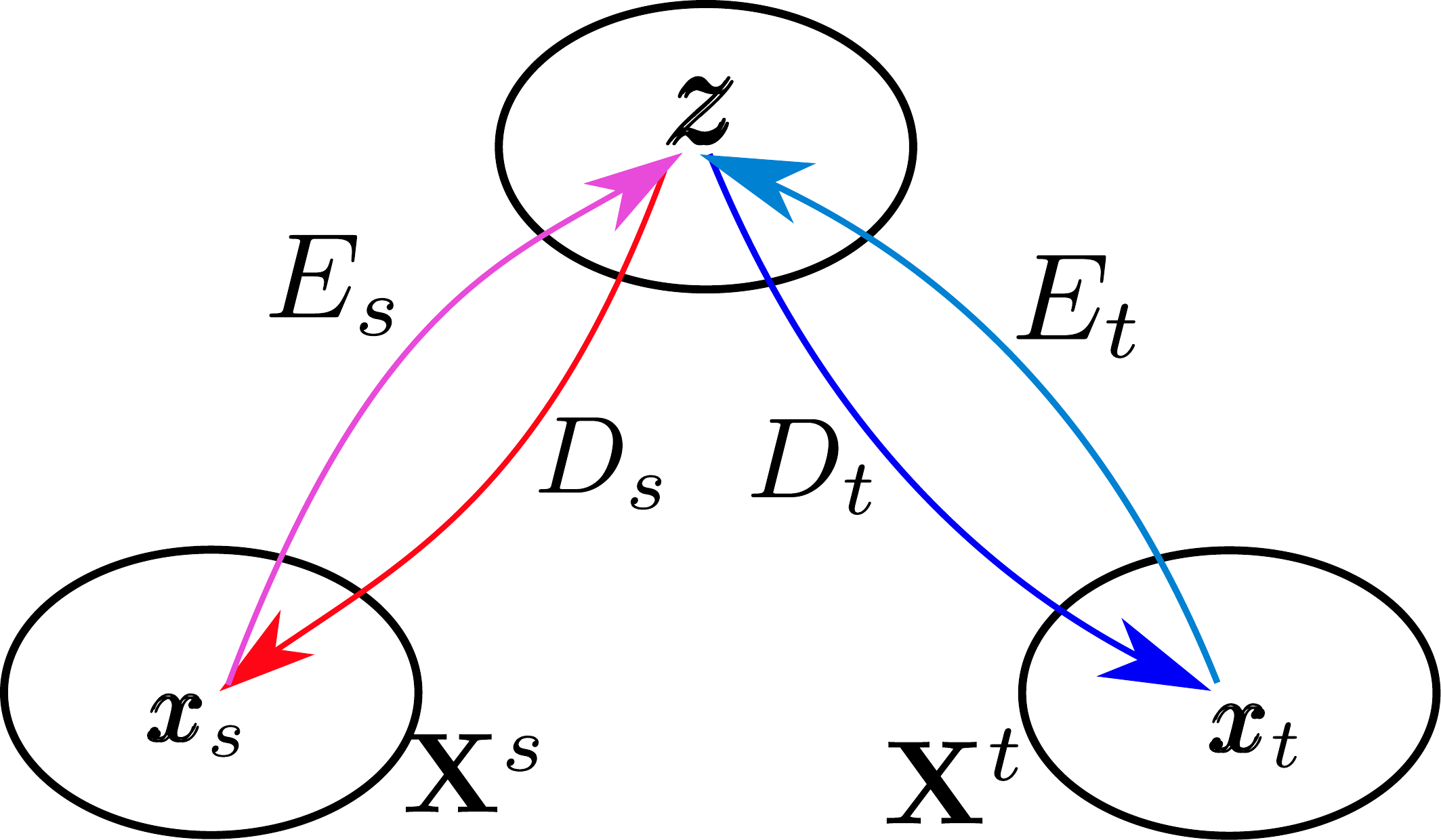}
        \caption{Shared latent space as assumed in UNIT. Images from different domains can be recovered in each domain or projected to another by latent codes in the shared latent space.}
        \label{fig:unit}
    \end{subfigure} \hfill
    \begin{subfigure}[t]{0.23\textwidth}
        \centering
        \includegraphics[width=0.9\textwidth]{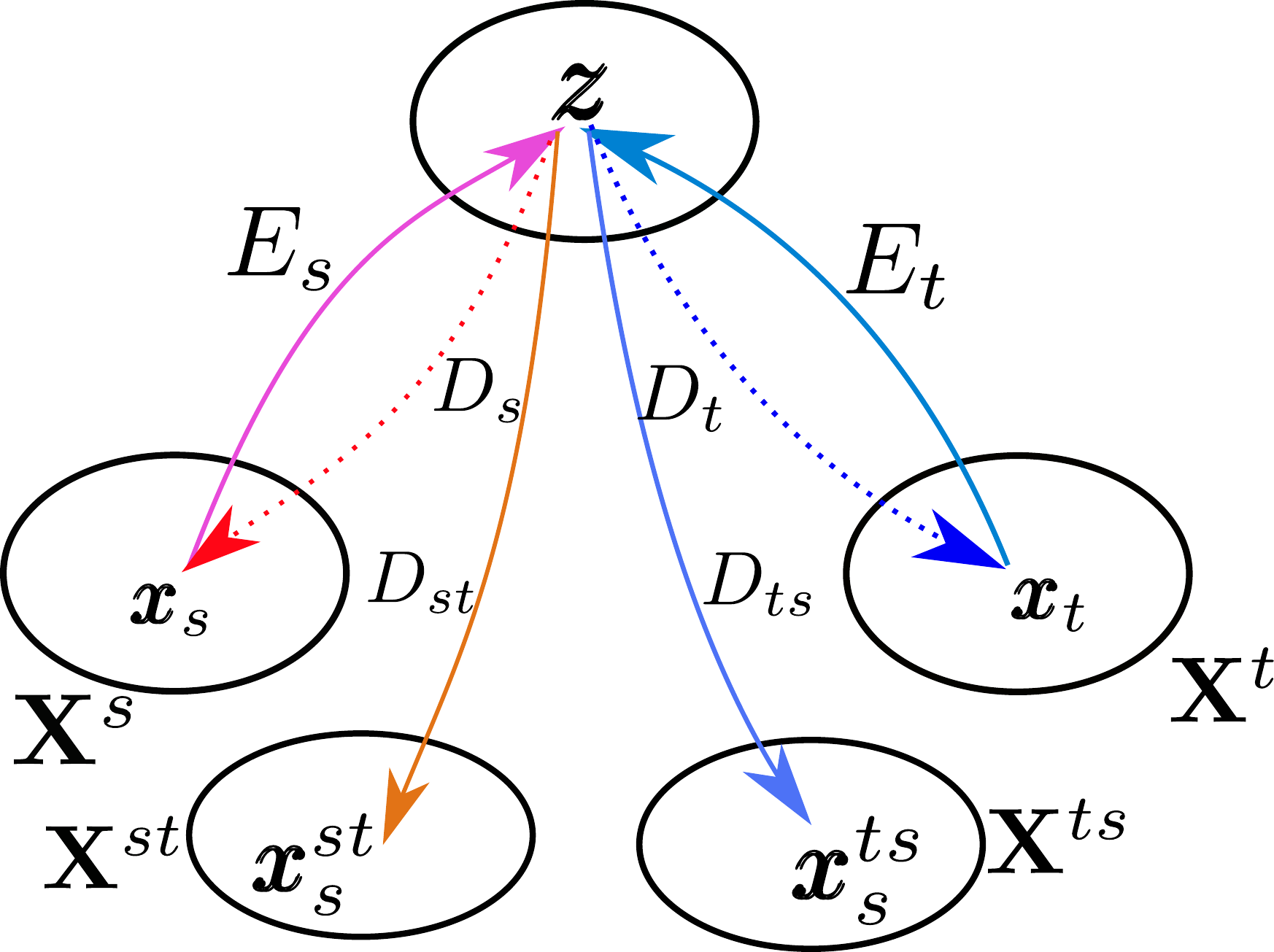}
        \caption{Proposed transition spaces based on a shared latent space. $D_{st}$ and $D_{ts}$ are cross-grafted decoders; $\bm {x}^{st}=D_{st}(\bm {z})$ and $\bm {x}^{ts}=D_{ts}(\bm {z})$ are the corresponding intermediate ``transitions''.}
        \label{fig:cgrs_sche}
    \end{subfigure} 
\end{figure}

From the probabilistic perspective, we would like to obtain a joint distribution for the source and target domain, then the marginal information of domains can be acquired. According to the coupling theory~\cite{Torgny2002Prob}, the marginal distribution could be obtained by an infinite set of joint distributions. To address this problem, a shared latent space is employed in UNIT. A pair of corresponding images $\bm x_s, \bm x_t$ in different domains can be mapped to the shared latent representation space $\bm z$ (Fig.~\ref{fig:unit}). Two parallel VAEs are then used for the shared latent representations extraction, using encoders ($E$) for latent encoding: $\bm z=E_s(\bm x_s)=E_t(\bm x_t)$, and decoders ($D$) for reconstruction: $\bm x_s=D_s(\bm z), \bm x_t=D_t(\bm z)$. After this, translation functions $F_{st}$ and $F_{ts}$ are learned to map the images from one domain to another, where $\bm {x_s}= F_{ts}(\bm {x}_t)=D_s(D_t(\bm {x}_t))$ and $\bm {x}_t= F_{st}(\bm {x}_s)=D_t(E_s(\bm {x}_s))$. The high level representations are shared between the encoders ($E_s,E_t$) and the decoders ($D_s, D_t$). 

\begin{figure*}[!h]
    \centering
    \includegraphics[width=0.8\textwidth]{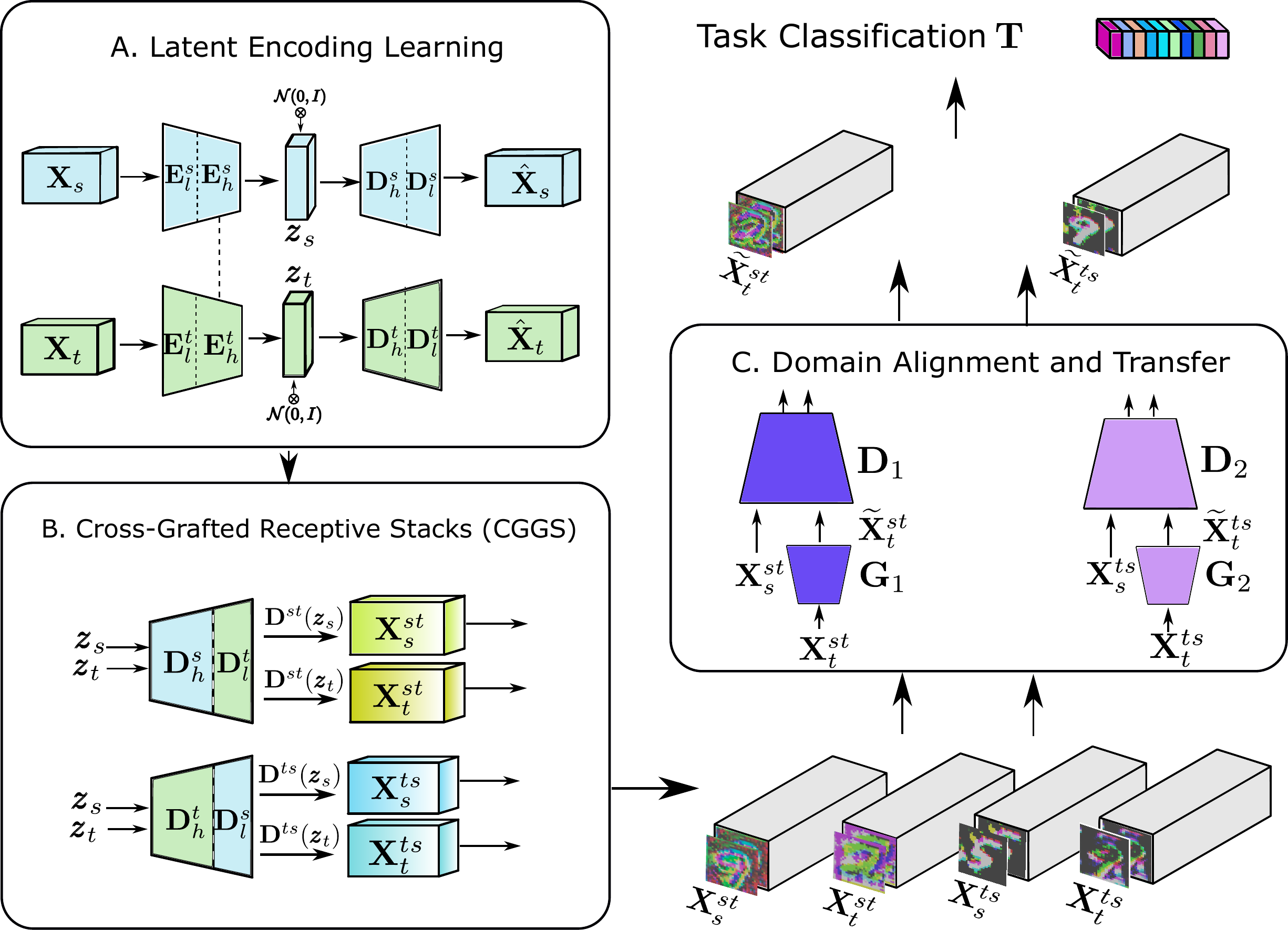}
    \caption{\small Overview of the the proposed model. Module \textit{A}: high-level layers of encoders $E_{h}^{s}$, $E_{h}^{t}$ are shared (indicated by the dashed line). Decoder layers $D_{h}^{s}$ and $D_{h}^{t}$ output high-level representations of the source and target images respectively, whereas $D_{l}^{s}$, $D_{l}^{t}$ give low-level representations;  Module B: $\mathbf{X}_s^{st}$, $\mathbf{X}_t^{ts}$, $\mathbf{X}_s^{ts}$, $\mathbf{X}_t^{ts}$ are the transitions reproduced by CGGS ($D^{st}\equiv \lbrack D_{h}^{s}\circ D_{l}^{t}\rbrack$ and $D^{ts}\equiv \lbrack D_{h}^{t}\circ D_{l}^{s}\rbrack$) from latent encodings; Module \textit{C}: $G_1$ and  $G_2$ are adversarial generators, while $D_1$, $D_2$ are discriminators. Best viewed in colors.}
    \label{fig:model_schem}
\end{figure*} 

Inspired by UNIT, we also assume a shared latent space for the joint latent representations. However, the latent encodings are not only decoded back to their own domains; rather, they are projected into some intermediate domain representations (Fig.~\ref{fig:cgrs_sche}), i.e., transition spaces. For transition generation, we relax the shared condition for the decoder, and construct cross-grafted generative stacks (CGGS). Specifically, the transition spaces can be constructed by cross-grafting the learned deep decoder stacks from the source and the target. According to the different grafted types in layers, it can generate various joint distribution spaces. UNIT could be viewed as a special case of our model. In our work, we exploit two special cases where the CGGS is constructed in a sequential way, say $D_{st}$ and $D_{ts}$. The outputs of mixed decoder (transitions), taking $D_{st}$ as example, are $\mathbf{X}_{s}^{st}=D_{st}(E_s(\mathbf{X}_s))$ and $ \mathbf{X}_{t}^{st}= D_{st}(E_t(\mathbf{X}_t))$.

Therefore, our motivation is to construct an intermediate pixel-level transition space with probability models $p(\mathbf {X_{st}})$ or $p(\mathbf {X_{ts}})$, so that samples from this space are similar to both domains and visually meaningful. We use CGGS to generate transition spaces based on latent encodings sampled from the shared latent space. These transitions can be seen as novel variations of images from the source and target domains, with reduced gap between transitions generated from the two domains. Adversarial learning in the GAN components is then used for label alignment and further to push the transitions from different domains to be closer.

\subsection{Model Description}

Our framework is illustrated in Figure~\ref{fig:model_schem}. There are three main modules in this end-to-end model as shown. Firstly, in module \textit{A}, a pair of VAEs are implemented by CNNs. Both the encoders and decoders are divided into high and low level stacks. The high-level layers of the encoders are shared between the domains. We assume that they have the same latent space with a normal prior $\mathcal{N}(0, I)$.

Secondly, the latent encodings pass through the cross-grafted stacks, forming cross-domain transitions. In module \textit{B}, we construct two parallel CGGS by grafting the decoder stacks of the source and the target. Therefore, the cross-domain transition images (${\mathbf{X}_s^{st}},{\mathbf{X}_t^{st}},{\mathbf{X}_s^{ts}},{\mathbf{X}_t^{ts}} )$ are generated when the latent encodings from different domains (indicated by subscripts) pass through the CGGS (order indicated by superscripts). 

Thirdly, in the domain alignment module \textit{C}, $G_1$ and $G_2$ are two adversarial generators for transitions. They are used to generate the target transition adversarial to the source's transition, and vice versa. The situation when the source-initiated transition works as the ``real player'' for the adversarial generation is shown in Figure~\ref{fig:model_schem}\footnote{The arrangement can be flexible, i.e. it also works if the target transition is used as the real player.}. Here the adversarial versions of the corresponding target-initiated transitions are $\widetilde{\mathbf{X}}_t^{st}$, and $\widetilde{\mathbf{X}}_t^{ts}$. The discriminators $D_1$, $D_2$ are used to distinguish transitions of  ${\mathbf{X}_s^{st}}$ from $\widetilde{\mathbf{X}}_t^{st}$, and ${\mathbf{X}_s^{ts}}$ from $\widetilde{\mathbf{X}}_t^{ts}$ respectively.  

Finally, the aligned transitions are fed into the task classifier to complete the adaptation. The training process adopts standard back-propagation. In contrast to the conventional domain adaptation framework in which the classifier input is $\{ \mathbf{X}_s, \pmb{y}_s\}$ and output is $\{ \mathbf{X}_t,\widehat{\pmb{y}_t}\}$, our model's classifier is trained by $\{\mathbf{X}_s^{st},\pmb{y}_s\}$, $\{\mathbf{X}_s^{ts},\pmb{y}_s\}$ and tested by $\{ \widetilde{\mathbf{X}}_t^{st},\pmb{y}_t\}$, $\{ \widetilde{\mathbf{X}}_t^{ts},\pmb{y}_t\}$. In short, the transitions of the source data are used for training, and the adversarial transitions of the target data  for testing.

\subsection{Adversarial generation of transitions}
\label{subsec:generation_cgrs}
In this section we present the adversarial generation of transitions from a probabilistic point of view, and show that the cross-domain transitions are perturbation spaces between the source and target domains. 

Firstly, we obtain the latent encodings of source and target domains using VAEs~\cite{Kingma2014}, assuming they have a normal prior distribution. They encode a data sample $\pmb{x}$ to a latent variable $\pmb{z}$ and decode it back to the data space with $\widehat{\pmb{x}}$. We obtain the latent encodings $\pmb{z}_s$ and $\pmb{z}_t$, which are conceptually sampled from conditional probability densities $q(\pmb{z}_s|\mathbf{X}_s)$ and $q(\pmb{z}_t|\mathbf{X}_t)$ respectively.

Cross-grafted generative stacks are then  constructed to map the encodings to the cross-domain transition spaces, which are later aligned to the source domain's transitions space by GAN. 
We explain the cross-grafting process in detail as follows. 

When the latent encoding $\pmb{z}_{k}$ ($\pmb{z}_{s}$ or $\pmb{z}_{t}$) passes through its original VAE, the generation of transitions can be expressed in a generative framework~\cite{Bengio2013}. 
Specifically, for $\pmb{z}_{s}$ to go through $D_s^h$, the higher stack of the decoder $D_s$, we have:
\begin{equation}\label{eq:high_level_transitions}
\textstyle \mathcal{P}_s^s=\textstyle 
\left(\prod_{c=1}^{N-1}p_{s}(\pmb{m}^{c+1}| 
\pmb{m}^{c},\theta_{D_s^h}^{c})\right)p_{s}(\pmb{m}^1|\pmb{z}_{s},
\theta_{D_s^h}^{1}),
\end{equation}
where $N$ is the number of high-level decoder layers, $\theta_{D_{ih}}^{c}$ denotes the network parameters of $D_s^h$ in the $c$-th layer, and  $\pmb{m}^{c}$ is the corresponding output space of that layer. Then, going through the lower stack of decoder $D_s^l$, $\pmb{m}^N$ is transformed to the final auto-association, with:
\begin{equation}\label{eq:low_level_transitions}
\textstyle \mathcal{P}_{s}^{ss}=\textstyle
\mathcal{P}_s^s
\left(\prod_{c=1}^{M-1}p_{s}(\pmb{n}^{c+1}| \pmb{n}^{c},\theta_{D_s^l}^{c})\right)p_{s}(\pmb{n}^1|\pmb{m}^{N},\theta_{D_s^l}^{1}),
\end{equation}
where $M$ is the number of low-level decoder layers, $\pmb{n}^{c}$ is the output space of low-level decoder of on layer $c$, and $\theta_{D_s^l}^{c}$ stands for the parameters of $D_s^l$ in that layer. 

Now, with CGGS, we first look at the route of $\pmb{z}_k\rightarrow D_h^s\rightarrow D_l^t$, i.e. the latent encodings $\pmb{z}_k$ from the source domain may go through the higher stack $D_h^s$ (from the source), followed by the lower stack $D_l^t$ (from the target), resulting in a transition representation  $\mathbf{X}_s^{st}$, which is related to the following generative model, with $D_l^t$ being interpreted as its counterpart $D_l^s$ added with perturbations $\pmb{\epsilon}_{tl}$ to all the lower layers: 
\begin{equation}\label{eq:low_level_recons_st}
\begin{array}{ll}
\textstyle
\mathcal{P}^{st}_s& =\displaystyle \mathcal{P}_s^s\left(\prod_{c=1}^{M-1}p_{t}(\pmb {n}^{c+1}| \pmb{n}^{c},\pmb{\theta}_{D_{l}^t}^{c})\right)p_{t}(\pmb {n}^1|\pmb{m}^{N},\pmb{\theta}_{D^t_{l}}^{1}) \\
& =\mathcal{P}_s^s\displaystyle \left(\prod_{c=1}^{M-1}p_{t}(\pmb {n}^{c+1}| \pmb{n}^{c},\pmb{\theta}_{D_{l}^s}^{c}+\pmb{\epsilon}_{tl}^c)\right)p_{t}(\pmb {n}^1|\pmb{m}^{N},\pmb{\theta}_{D^s_{l}}^{1}+\pmb{\epsilon}_{tl}^1).
\end{array}
\end{equation}

Therefore transition  $\mathcal{P}^{st}_s$ can be seen as a perturbed version of the reconstruction $\mathcal{P}_s^{ss}$. Similarly, with another route $\pmb{z}_k\rightarrow D_h^t\rightarrow D_l^s$,  source latent encodings $\pmb{z}_s$ first go through the high-level stack $D_h^t$, which is in effect a perturbation from $D_h^s$ (tilde symbols indicate new output counterparts):  
\begin{equation}\label{eq:high_level_pert}
\begin{array}{ll}
\mathcal{P}^t_s & = \left(\displaystyle \prod_{c=1}^{N-1}p_{t}(\tilde{\pmb{m}}^{c+1}| \tilde{\pmb{m}}^c,\pmb{\theta}_{D_{h}^t}^{c})\right) p_{t}(\tilde{\pmb{m}}^1|\pmb{z}_{s}, \pmb{\theta}_{D_{h}^t}^{1}) \\
 & =\displaystyle\left(\prod_{c=1}^{N-1}p_{t}(\tilde{\pmb{m}}^{c+1}| \tilde{\pmb{m}}^c,\pmb{\theta}_{D_{h}^s}^{c}+\pmb{\epsilon}_{th}^c)\right) p_{t}(\tilde{\pmb {m}}^1|\pmb{z}_{s}, \pmb{\theta}_{D_{h}^s}^{1}+\pmb{\epsilon}_{th}^1),
\end{array}
\end{equation}
and the perturbed output $\widetilde{\pmb{m}}^N$ will go through $D_l^s$, the lower stack from the source's decoder:
\begin{equation}\label{eq:assoc_ts}
\textstyle \mathcal{P}^{ts}_s  =\displaystyle \mathcal{P}_s^t\left(\prod_{c=1}^{M-1}p_{s}(\tilde{\pmb{n}}^{c+1}|\tilde{\pmb{n}}^{c},\pmb{\theta}_{D_{l}^s}^{c})\right)p_{s}(\tilde{\pmb {n}}^1|\widetilde{\pmb{m}}^{N},\pmb{\theta}_{D^s_{l}}^{1}). 
\end{equation}
Again, we see that $\mathcal{P}^{ts}_s$ is a perturbed version of $\mathcal{P}^{ss}_s$. The other transitions $\mathcal{P}^{st}_t$ and $\mathcal{P}^{ts}_t$ can be similarly analyzed as being the perturbed versions of $\mathcal{P}^{tt}_t$. These transitions bridge the gap between the source and target domains. 

The transitions are constructed, but the alignment of transitions to the label space of the source domain is yet to be done. For example, we can easily migrate the label information from source $\mathbf{X}_s$ to the source-initiated transition $\mathbf{X}_s^{st}$. Yet for the target-initiated transition $\mathbf{X}_t^{st}$, this is not straightforward, as the target pattern $x_t$ is not paired with the source pattern $x_s$ with label consistency --- rather, $x_t$ can be an arbitrary target pattern. To get the label distributions aligned, we use generator and discriminator pairs ($D_1$ and $G_1$; $D_2$ and $G_2$) to align the transitions generated by different latent encodings. Following the discussion given in~\cite{Goodfellow2014}, the adversarial game between the generators and the discriminators can be seen as a learning process in effect to minimize the Jensen-Shannon divergence (JSD) between the transitions and their generative counterparts:
\begin{equation}\label{eq:jsd1}
\begin{gathered}
\textstyle \widetilde{\mathbf{X}}_t^{ij} \sim p(\widetilde{\mathbf{X}}_t^{ij}|\mathbf{X}_t^{ij},\theta_D,\theta_{G_{k}}) \\
\textstyle w.r.t~\min \mathrm{JSD}\left(p(\mathbf{X}_{s}^{ij})\|p(\widetilde{\mathbf{X}}_{t}^{ij})\right),
\end{gathered}
\end{equation}
where $\theta_{G_{k}}$ ($G_{1}$ when $i=s, j=t$ and $G_{2}$ for $i=t, j=s$) are used as generators for $\widetilde{\mathbf{X}}_{t}^{st}$ and $\widetilde{\mathbf{X}}_{t}^{ts}$ during the alignment, as shown in Figure~\ref{fig:model_schem}. 
The alignment process reduces the domain gap between data being used for training and testing (e.g. $\mathbf{X}_s^{st}$ and $\widetilde{\mathbf{X}}_{t}^{st}$), leading to more robust transfer learning perfomance.


\subsection{Learning in DATL}
\label{subsec:learning}

To train our model, we jointly solve the learning problems of all its modules. Our loss function contains four parts: loss for the within-domain VAEs~\cite{Kingma2014}, loss for the cross-domain adversarial learning, content similarity, and finally, the classifier training loss.

First, we need to learn the representations of the source and target domains from the encoders and decoders. Here, we minimize the within-domain VAEs loss functions, which contain both the reconstruction error and a prior regularization: 
\begin{equation} \label{eq:overall_vae}
L_{\mathit{VAEs}}=\lambda_1L_{rec}+\lambda_2L_{prior},
\end{equation}
where the reconstruction error is given by
\begin{equation} \label{eq:vae_rec}
\begin{split}
\textstyle L_{rec}  &= \textstyle -\{ \mathbb{E}_{q_{s}(\pmb{z}_s|\mathbf{X}_s)}[\log p_{s}(\mathbf{X}_s|\pmb{z}_s)] \\
\textstyle &+ \mathbb{E}_{q_{t}(\pmb{z}_t|\mathbf{X}_t)}[\log p_{t}(\mathbf{X}_t|\pmb{z}_t)]\} ,\\
\end{split}
\end{equation}
and the prior regulation is on minimizing the 
Kullback-Leibler divergence (KLD) between the prior and the variational encoding priors: 
\begin{equation} \label{eq:vae_latent}
\textstyle L_{prior} 
= \textstyle  \mbox{KLD}(q_{s}(\pmb{z}_s|\pmb{x}_{s})||p(\pmb{z})) 
 + \mbox{KLD}(q_{t}(\pmb{z}_t|\pmb{x}_{t})||p(\pmb{z})),
\end{equation}
and $\lambda_1$ and $\lambda_2$ are the weights on the reconstruction and the variational encoding parts respectively. 

To align the source and target domains, we use the adversarial learning for the two transition spaces $\mathcal{P}_{st}$ and $\mathcal{P}_{ts}$. Specifically, for the source-initiated transition $\mathbf{X}_s^{st}$, a generative adversarial, $\tilde{\mathbf{X}}_t^{st}$, is learned based on the target-initiated transition $\mathbf{X}_t^{st}$. The adversarial objectives $L_G^{st}$ is: 
\begin{equation} \label{eq:gan_st}
\begin{split}
\textstyle L_G^{st} &= \textstyle 
\mathbb{E}_{x_{s},z_{s}}[\log D_{1}(\mathbf{X}_s^{st})] \\
\textstyle &+ \mathbb{E}_{x_{t},z_{t}}[\log (1- 
D_{1}(G_1(\mathbf{X}_t^{st}))],
\end{split}
\end{equation}
where 
$D_1(x)$ is the probability function assigned by the discriminator network, which tries to distinguish the generated source-based transitions from the target-based ones. For $\mathcal{P}_{ts}$, the adversarial objective $L_G^{ts}$ is similarly defined:
\begin{equation} \label{eq:gan_ts}
\begin{split}
\textstyle L_G^{ts} &= \textstyle 
\mathbb{E}_{x_{s},z_{s}}[\log D_{2}(\mathbf{X}_s^{ts})] \\
\textstyle &+ \mathbb{E}_{x_{t},z_{t}}[\log (1- 
D_{2}(G_2(\mathbf{X}_t^{ts}))]
.
\end{split}
\end{equation}
The overall adversarial generative cost function is therefore a sum weighted by a hyperparameter $\lambda_0$: 
\begin{equation} \label{eq:overall_gan}
\textstyle L_{G}=\lambda_0(\textstyle L_G^{st}+L_G^{ts}).
\end{equation}
For training stability, we introduce a content similarity metric for the transitions~\cite{Bousmalis2017, Kim2017}. Either $L1$ or $\mathnormal{L}2$ penalty can be used to regularize the transitions, such as MSE, pairwise MSE, and Huber loss. Here we simply use MSE. The MSE loss for transitions $\mathbf{X}^{st}$ is given as follows:
\begin{equation}
\textstyle L_c^{st} = \textstyle 
\mathbb{E}_{(\mathbf{X}_s,\mathbf{X}_t)}||\mathbf{X}_s^{st} - 
\mathbf{X}_t^{st}||_2^{2}.
\end{equation}
For $\mathbf{X}^{ts}$, $L_s^{ts}$ is similarly defined. The overall 
content objective for transitions is weighted by parameter $\lambda_3$:
\begin{equation} \label{eq:overall_pmse}
\textstyle L_c = \lambda_3(L_c^{st} + L_c^{ts}).
\end{equation}
For the final loss component, we use a typical soft-max cross-entropy loss for the classification task:
\begin{equation} \label{eq:classifier}
\textstyle L_{T} = \mathbb{E}_{\mathbf{X}_s}[-\mathbf{y}_{s}^{T}\log T_s(\mathbf{X}_s^{st}) - \mathbf{y}_{s}^{T}\log T_s(\mathbf{X}_s^{ts})],
\end{equation}
where $\mathbf{y}_{s}$ is the class label for source $\mathbf{X}_{s}$, and $T_s$ is the task classifier. 

Finally, the overall loss function of our model is:
\begin{equation} \label{eq:overall_loss_function}
L^*
=\mathop{\min}\limits_{G,E,D,T}\mathop{\max}\limits_{D_1, D_2}
(L_{\mathit{VAEs}}+ L_{G}+L_{c}+L_{T}).
\end{equation}

\begin{algorithm}[!h]
    \SetAlgoLined 
    \DontPrintSemicolon
    \KwIn{Source: $\mathbf{X}_s, \pmb{y}_s$, Target:$\mathbf{X}_t$}
    \KwResult{Model Weights of DATL}
    $\theta_E, \theta_D, \theta_G, \theta_{Dis}, \theta_T$ $\leftarrow$ 
    initialization\;
    \For{iterations of traning}{
        $\pmb{z}$ $\leftarrow$ sample from $\mathcal{N}(0, I)$ \\
        $\mathbf{X}_s, \mathbf{X}_t$ $\leftarrow$ sample mini-batch \\
        $\pmb{z}_s, \pmb{z}_t$ $\leftarrow$ sample from $q(\pmb{z}_s|\mathbf{X}_s)$ 
        and $q(\pmb{z}_t|\mathbf{X}_t)$ \\
        
        Generate ${\mathbf{X}_s^{st}},{\mathbf{X}_t^{st}}, {\mathbf{X}_s^{ts}},
        {\mathbf{X}_t^{ts}}$ by feeding $\pmb{z}_s, \pmb{z}_t$ through CGGS \\ 
        Generate $\widetilde{\mathbf{X}}_t^{st}$,  $\widetilde{\mathbf{X}}_t^{ts}$ using $(G_1, D_1)$ and $(G_2, D_2)$\\
        $L_{\mathit{VAEs}}, L_{G}, L_{s}, L_{T}$ $\leftarrow$ caculated by 
        (\ref{eq:overall_vae}) to (\ref{eq:classifier}) \\
        \For{iterations of VAE updating}{
            $\theta_E, \theta_D$ $\leftarrow$ --$\Delta_{\theta_E, \theta_D}$
            ($L_{\mathit{VAEs}}$) \\
        }
        \For{iterations of discriminator updating}{
            $\theta_{Dis}, \theta_T$ $\leftarrow$ --$\Delta_{\theta_{Dis}, 
                \theta_T}$($L_G+L_T$) \\
        }
        \For{iterations of generator updating}{
            $\theta_G$ $\leftarrow$ --$\Delta_{\theta_G, 
                \theta_E}$($L_G+L_c+L_T$) \\
        }
    }
    \caption{Training of the DATL framework}
\end{algorithm}  

We solve this minimax problem of the loss function optimization by three alternating steps. First, the latent encodings are learned by the self-mapped process, which updates $(E_s, E_t, D_s, D_t) $, but keeps CGGS $(D^{st},D^{ts})$, Discriminators ($D_1, D_2$), and Generators ($G_1, G_2$) fixed. Then, we update $D_1$, $D_2$ {and the classifier} $T$, while keeping the two VAE channels, CGGS stacks, and the generators fixed. Finally, we update $(E_1, E_2, G_1, G_2)$,  while 
all other components are fixed.

\begin{table}[]
    \centering
    \resizebox{\columnwidth}{!}{
        \begin{tabular}{|l|c|c|c|c|}
            \hline
            \small Models & \makecell{\small Pixel-level \\Generation} & \makecell{\small Relaxed Consis-\\tency Regularization} & \makecell{\small New adapta-\\tion Space} & \makecell{\small Flexible\\ Generation} \\
            \hline\hline
            \small PixelDA~\cite{Bousmalis2017} &  \Pisymbol{pzd}{51} & \Pisymbol{pzd}{51}  & \Pisymbol{pzd}{55}  & \Pisymbol{pzd}{55}    \\
            \small UNIT~\cite{Liu2017} &  \Pisymbol{pzd}{51} & \Pisymbol{pzd}{55}  & \Pisymbol{pzd}{55}  & \Pisymbol{pzd}{55}    \\
            \small CyCADA~\cite{Hoffman2018} &  \Pisymbol{pzd}{51} & \Pisymbol{pzd}{55}  & \Pisymbol{pzd}{55}  & \Pisymbol{pzd}{55}    \\
            \small GtA~\cite{Sankaranarayanan2018} &  \Pisymbol{pzd}{51} & \Pisymbol{pzd}{51}  & \Pisymbol{pzd}{55}  & \Pisymbol{pzd}{55}    \\
            \small DATL &  \Pisymbol{pzd}{51}  & \Pisymbol{pzd}{51}  & \Pisymbol{pzd}{51}  & \Pisymbol{pzd}{51}    \\
            \hline
    \end{tabular}}
    \caption{\small Properties comparison of recent pixel-level UDA models.}
    \label{table:properties_comparison}
\end{table}

\begin{figure*}
    \centering
    \includegraphics[width=\textwidth]{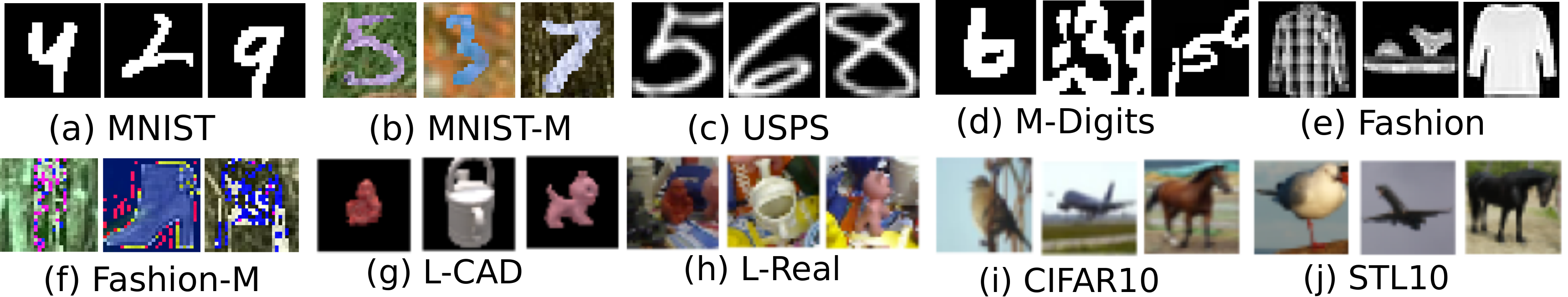}
    \caption{\small Example images from the datasets used in our experiments.}
    \label{fig:datasets}
\end{figure*}  

We compare the properties of the DATL model with several state-of-the-art generative models for unsupervised domain adaptation in Table~\ref{table:properties_comparison}. Compared with the feature-level unsupervised domain adaptation methods, generative pixel-level models can obtain visually compelling results. For some of these adaptation methods such as UNIT and CyCADA, a cycle-consistency loss is necessary in order to ensure the efficient transformation between two domains, which increases the complexity of the training process. 
In contrast, our model avoids the use of cycle-consistency but relies on an effective combination of content similarity and adversarial learning to make the generated transition images resemble both those of both domains. 
Our model can also be flexibly customized for the given tasks, i.e. the exact setting of CGGS can be tuned depending on the adaptation needs.

\section{Experiments} \label{sec:experiments}


\begin{table*}[thbp]
    \centering
    \resizebox{1.0\textwidth}{!}{
        \begin{tabular}{|l|c|c|c|c|c|c|c|c|}
            \hline
            Source & MNIST & MNIST-M & MNIST & USPS & MNIST  & M-Digits & Fashion & Fashion-M \\
            Target & MNIST-M & MNIST & USPS  & MNIST & M-Digits & MNIST & Fashion-M & Fashion \\
            \hline\hline
            Source Only & 0.561  &  0.633   & 0.634 & 0.625  & 0.603 & 0.651 & 0.527 & 0.612   \\
            \hline\hline
            CyCADA~\cite{Hoffman2018} & 0.921 & 0.943 & 0.956  & 0.965  & 0.912 & 0.920 & 0.874 & 0.915 \\
            GtA~\cite{Sankaranarayanan2018} & 0.917 & 0.932 & 0.953  & 0.908  & 0.915 & 0.906 & 0.855 & 0.893 \\
            CDAN~\cite{Long2018} & 0.862 & 0.902 & 0.956 & \textbf{0.980} & 0.910 & 0.922 & 0.875 & 0.891 \\
            DANN~\cite{Ganin2016} & 0.766 & 0.851  & 0.774 & 0.833   & 0.864  & 0.914  & 0.604  & 0.822  \\
            PixelDA~\cite{Bousmalis2017} & 0.982 & 0.922   & 0.959 & 0.942   & 0.734 & 0.913  & 0.805   &  0.762  \\
            UNIT~\cite{Liu2017} & 0.920 &  0.932 &  0.960 & 0.951   & 0.903  & 0.910  & 0.796   &  0.805  \\
            \hline\hline
            DATL ($\mathbf{X}^{st}$) &  0.890 & \textbf{0.983}  & \textbf{0.961}  & 0.956  & \textbf{0.916} & \textbf{0.923}  & 0.853 & \textbf{0.917}     \\
            DATL ($\mathbf{X}^{ts}$) & \textbf{0.983}  &  0.871   & 0.943  &  0.953  & 0.883  & 0.892  & \textbf{0.886} & 0.903     \\
            \hline\hline
            Target Only & 0.983  & 0.985  &0.980 &  0.985   & 0.982  & 0.985  & 0.920  &  0.942   \\
            \hline      
    \end{tabular}}
    \caption{\small Mean classification accuracy comparison. The "source only" row is the accuracy for target without domain adaptation training only on the source. And the "target only" is the accuracy of the full adaptation training on the target. For each source-target task the best performance is in bold.}
    \label{table:accuracy}
\end{table*}

We have evaluated our model on some benchmark datasets used commonly in the domain adaptation literature for a number of different adaptation scenarios, including digits and non-digits, synthetic and real images, and multi-class object images. 
Details of these datasets will follow. Example images of these datasets are shown in Figure~\ref{fig:datasets}.
We compare our DATL method with the following state-of-the-art domain adaptation methods: Pixel-level domain adaptation (PixelDA)~\cite{Bousmalis2017}, Domain Adversarial Neural Network (DANN)~\cite{Ganin2016}, Unsupervised Image-to-Image translation (UNIT)~\cite{Liu2017}, Cycle-Consistent Adversarial Domain Adaption (CyCADA)~\cite{Hoffman2018}, and ``Generate to Adapt'' (GtA)~\cite{Sankaranarayanan2018}. 

In addition, we also use source-only and target-only training as reference scenarios to provide the performance lower- and upper-bounds respectively, following the practice in~\cite{Bousmalis2017, Ganin2016}. For source-only training, the model is trained on the source dataset only, and then  tested on the target dataset. For target-only, the target dataset is used for training and testing. 

\subsection{Datasets and Adaptation Tasks}
\label{subsec:datasets}

We use ten datasets and construct six domain adaptation tasks, all with  bidirectional adaptation considered except the Linemod task.

\noindent
\textbf{MNIST $\rightleftarrows$ MNIST-M:} This is a scenario when the image content is the same, but the target data are polluted by noise. MNIST handwritten dataset~\cite{LeCun1998} is a very popular machine learning dataset. It has a training set of 60,000 binary images, and a test set of 10,000. There are 10 classes in the dataset. MNIST-M~\cite{Ganin2016} is a modified version for the MNIST, with random RGB background cropped from the Berkeley Segmentation Dataset\footnote{URL https://www2.eecs.berkeley.edu/Research/Projects/CS/vision/bsds/} inserted. In our experiments, we use the standard split of the dataset.

\noindent
\textbf{MNIST $\rightleftarrows$ USPS:} For this task, the source and target domains have different contents but the same background. USPS is a handwritten ZIP digits dataset~\cite{LeCun1989}. It is collected by the U.S. Postal Service from envelopes processed at the Buffalo, N.Y Post Office. It contains 9298 binary images ($16\times16$), 7291 of which are used for training, the remaining 2007 for testing. The USPS samples are resized to $28\times28$, the same as MNIST.   

\noindent
\textbf{Fashion $\rightleftarrows$ Fashion-M}: Fashion-MNIST~\cite{Xiao2017} contains 60,000 images for training, and 10,000 for testing. All the images are of grayscale, $28\times28$ in size. The samples are collected from 10 fashion categories: T-shirt/Top, Trouser, Pullover, Dress, Coat, Sandal, Shirt, Sneaker, Bag and Ankle Boot. There are some complex textures in the images. Following the protocol in~\cite{Ganin2016}, we add random noise to the Fashion images to generate the Fashion-M dataset as the counterpart.

\noindent
\textbf{MNIST $\rightleftarrows$ M-Digits}: In this task, we designed a multi-digit dataset to evaluate the proposed model, noted as ``M-Digits''. The MNIST digits are cropped first, and then randomly selected, combined and randomly aligned in a new image, limited to 3 digits in maximum. The label for the new image is decided by the central digit. All images are resized to $28\times28$. 

\noindent
\textbf{CIFAR10 $\rightleftarrows$ STL10}:  CIFAR10 is a labeled subset of a dataset with 80-million tiny images. It has 10 classes, and each class has 5000 training images and 1000 testing images, all of size 32$\times$32. The other dataset, STL10, also has 10 classes, and the images are acquired from the labeled ImageNet data. There are 500 training images and 800 testing images for each class. The two datasets share 8 common classes: airplane, bird, cat, deer, dog, horse, ship, truck. These classes are used for the adaptation task. Images in STL10 are resized to 32$\times$32.

\noindent
\textbf{Linemod 3D synthetic $\rightarrow$ Real images}: We followed the protocol of ~\cite{Bousmalis2017} and rendered the LineMod~\cite{Hinterstoisser2012} dataset for the adaptation between synthetic 3D images (source) and real images (target). We note them as L-CAD and L-Real in short. The objects with different poses are located at the center of the images; the synthetic 3D models are rendered with a black background, while the real images feature a variety of complex indoor scenes. Only the RGB images are used, the depth images are not. We use the standard cropped version~\cite{Wohlhart2015}. There are 16962 synthetic 3D images and 13328 real images.  

\subsection{Implementation Details}
\label{sec:implementation}

All the models are implemented using TensorFlow~\cite{Abadi2016} and are trained with mini-batch gradient descent using the Adam optimizer~\cite{Kingma2014c}. The initial learning rate is 0.0002. Then it adopts an annealing method, with a decay of 0.95 after every 20,000 mini-batch steps. The mini-batch size for both the source and target domains are 64 samples, and the input images are rescaled to [-1, 1]. The hyper-parameters for the loss function are $\lambda_0 = 1$, $\lambda_1=10$, $\lambda_2=0.01$, $\lambda_3=1$.   

In our implementation, the latent space is sampled from a normal distribution $\mathcal{N}(0, I)$, and is achieved by the convolutional encoders. The transpose convolution~\cite{Zeiler2010} is used in the decoder to build the reconstruction image space. This follows a similar structure as given in~\cite{Liu2017}, but we modified the padding strategy to `same' for convolution layers. For sake of convenience in the experiments, we added another 32-kernel layer before the last layer in the decoders. The stride is 2 for down-sampling in the encoders, and their counterpart in decoders is also 2 so as to get the same dimensionality of the original image. The encoders for the source and target domains share their high-level layers. We added the batch normalization between each layer in the encoders and the decoders. The stride step remains 1 for all the dimensions in the adversarial generators, and the kernel is $3\times3$. This adopts the structure of PixelDA~\cite{Bousmalis2017}, which uses a ResNet architecture. The discriminators fuse the domains, and are also part of the task classifier for  label space alignment. These follow the design in~\cite{Liu2017}. However, we do not share the  discriminators layers of $\mathbf{X}^{st}$ and $\mathbf{X}^{ts}$ channels. Also, we replace the max-pooling with a stride of $2\times2$ steps.

\subsection{Results}
\subsubsection{Quantitative Results}

Now we report the classification performance of our proposed model. During the experiments, transitions $\mathbf{X}_s^{st}$ and $\mathbf{X}_s^{ts}$ are used to train the classifier, and the adversarial generation of $\mathbf{X}_t^{st}$ and $\mathbf{X}_t^{ts}$ are used for testing. For the various adaptation tasks, we cite the performance from the literature where applicable, otherwise the performance are obtained by using the open-source code provided by the relevant work with the suggested optimal parameters. 

The accuracy of the target domain classification after domain adaptation is listed in Table~\ref{table:accuracy}, with results of 10 methods (2 versions of our model DATL, 6 state-of-the-art methods, plus the lower and upper bounds) across 4 tasks (each in two adaptation directions). Our proposed model outperforms the state-of-the-art in most of the scenarios, especially when content similarity is considered. Also, it can be seen that the adaptation performance is often asymmetric for the methods in comparison, e.g.~the accuracies for MNIST$\rightarrow$M-Digits and M-Digits$\rightarrow$MNIST are quite different for DANN and PixelDA. The DATL models, however, perform almost equally well on both directions for these adaptation tasks. 

\begin{figure}[thbp]
    \centering
    \includegraphics[width=0.8\columnwidth]{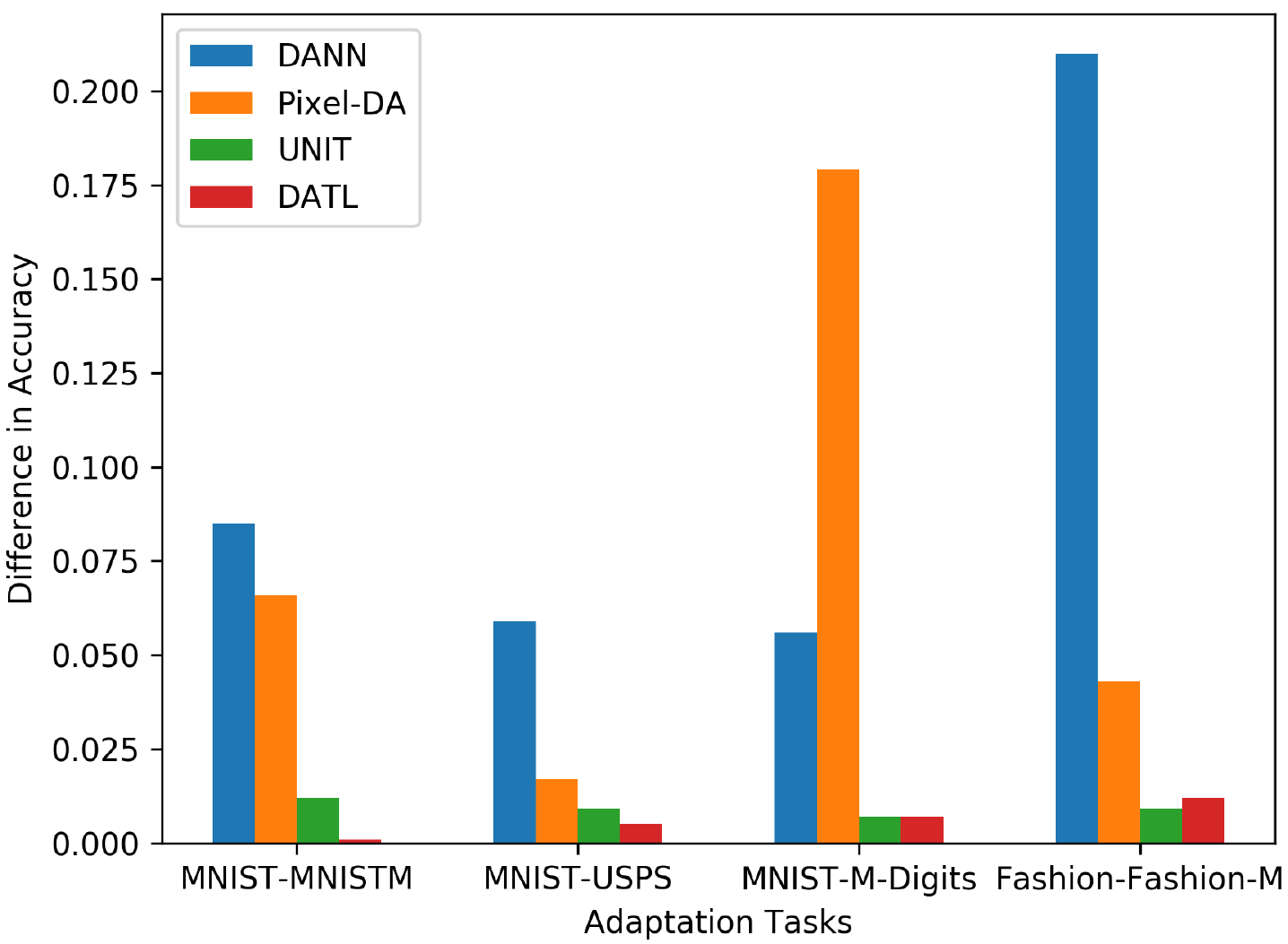}
    \caption{Absolute accuracy difference between different directions of adaptation for four tasks.}
    \label{fig:bidir_diff}  
\end{figure}

For MNIST$\rightleftarrows$MNIST-M and MNIST$\rightleftarrows$USPS, the mean classification accuracy nearly reaches the upper bound, suggesting these are easier tasks. On the other hand, we can see the adaptation task between Fashion and Fashion-M is more difficult than others. For this task, our method again not only achieves the best performance but also demonstrates balanced performance in two directions, as shown in Fig~\ref{fig:bidir_diff}, where the absolute accuracy difference between two directions of each adaptation task is given. 

To evaluate the effectiveness of our model to adapt from synthetic data to real images, we use the rendered Linemod dataset as described in Section \ref{subsec:datasets}. 
The benchmarks were cited from~\cite{Bousmalis2017}. From the results, we can see that the DATL achieved the best adaptation performance comparable to PixelDA. However, it consumes less samples to get the reasonable adaptation performance.

\begin{table}[htbp]  
    \centering
        \begin{tabular}{|c|c|}
            \hline
            Source &  L-CAD  \\
            Target & L-Real  \\
            \hline\hline
            Source-Only  & 0.447  \\
            \hline
            MMD~\cite{Tzeng2014} & 0.723   \\
            PixelDA~\cite{Bousmalis2017} & \textbf{0.998}    \\
            \hline 
            DATL ($\mathbf{X}^{st}$) & \textbf{0.998}  \\
            DATL ($\mathbf{X}^{ts}$) & 0.990  \\
            \hline
            Target-Only & 0.998 \\
            \hline
        \end{tabular}
    \caption{\small Performance comparison on the LineMod CAD to real images adaptation task.}
    \label{table:linemod}
\end{table}	

Finally, the adaptation performance of the CIFAR10$\leftrightarrow$ STL-10 task is given in Table \ref{table:linemod}. Compared with other tasks, this task seems to be more challenging, with an increased  domain gap to be tackled. 
Although in general the adaptation accuracy is much worse compared with the other tasks, DATL achieves the best performance on both adaptation directions, with the least directional difference. 

\begin{table}[thbp]
    \centering
        \begin{tabular}{|c|c|c|}
            \hline
            Source &  CIFAR-10 & STL-10  \\
            Target & STL-10 & CIFAR-10   \\
            \hline\hline
            Source-Only  & 0.541 & 0.636   \\
            \hline
            DANN~\cite{Ganin2016} & 0.569  &  0.661   \\
            DRCN~\cite{Ghifary2016} & 0.588  & 0.663  \\
            \hline 
            DATL ($\mathbf{X}^{st}$) & 0.613 & \textbf{0.672}    \\
            DATL ($\mathbf{X}^{ts}$) & \textbf{0.615} & 0.652   \\
            \hline
            Target-Only & 0.791 & 0.766   \\
            \hline
        \end{tabular}
    \caption{\small Adaptation performance for the STL-10 $\leftrightarrow$ CIFAR10 tasks.}
    \label{table:cifar} 
\end{table}	

\subsubsection{Qualitative Results}
\label{subsubsec:qualitative_results}

\begin{figure*}[thbp]
    \centering
    \begin{subfigure}[t]{0.48\textwidth}
        \centering
        \includegraphics[width=1.0\textwidth]{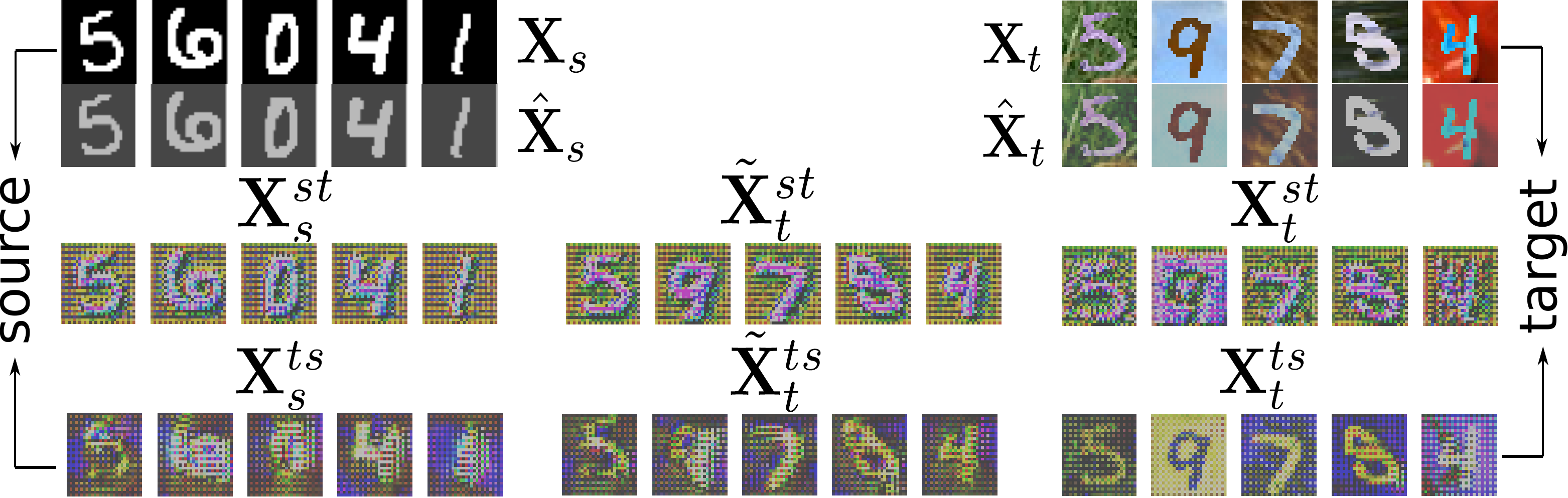}
        \caption{\small MNIST $\rightarrow$ MNIST-M}
        \label{fig:mnist_mnistm}
    \end{subfigure}
    \begin{subfigure}[t]{0.48\textwidth}
        \centering
        \includegraphics[width=1.0\textwidth]{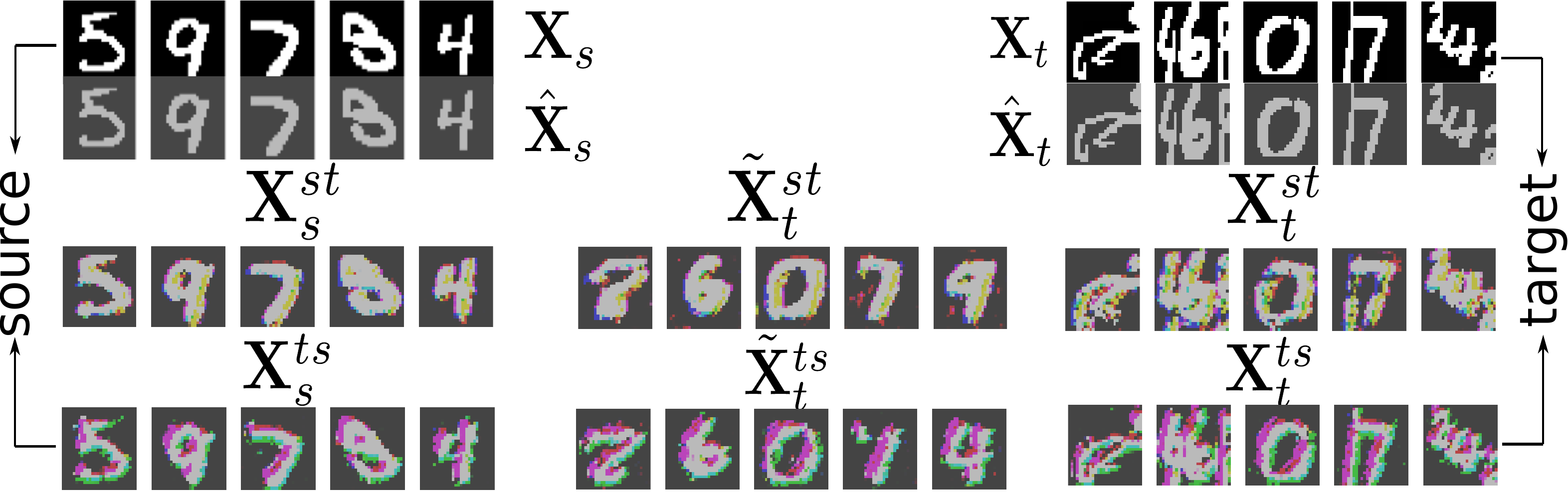}
        \caption{\small MNIST $\rightarrow$ M-Digits}
        \label{fig:mnist_M-digits}
    \end{subfigure}
    \begin{subfigure}[t]{0.48\textwidth}
        \centering
        \includegraphics[width=1.0\textwidth]{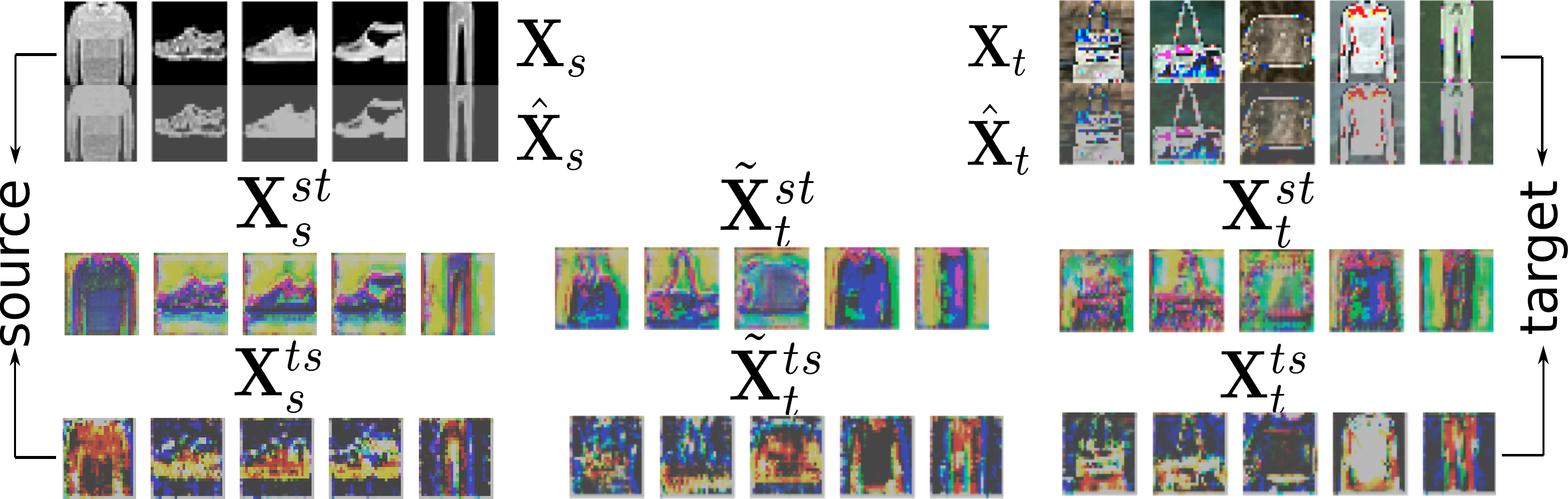}
        \caption{\small Fashion $\rightarrow$ Fashion-M}
        \label{fig:fashion-fashionm}
    \end{subfigure}
    \begin{subfigure}[t]{0.48\textwidth}
        \centering
        \includegraphics[width=1.0\textwidth]{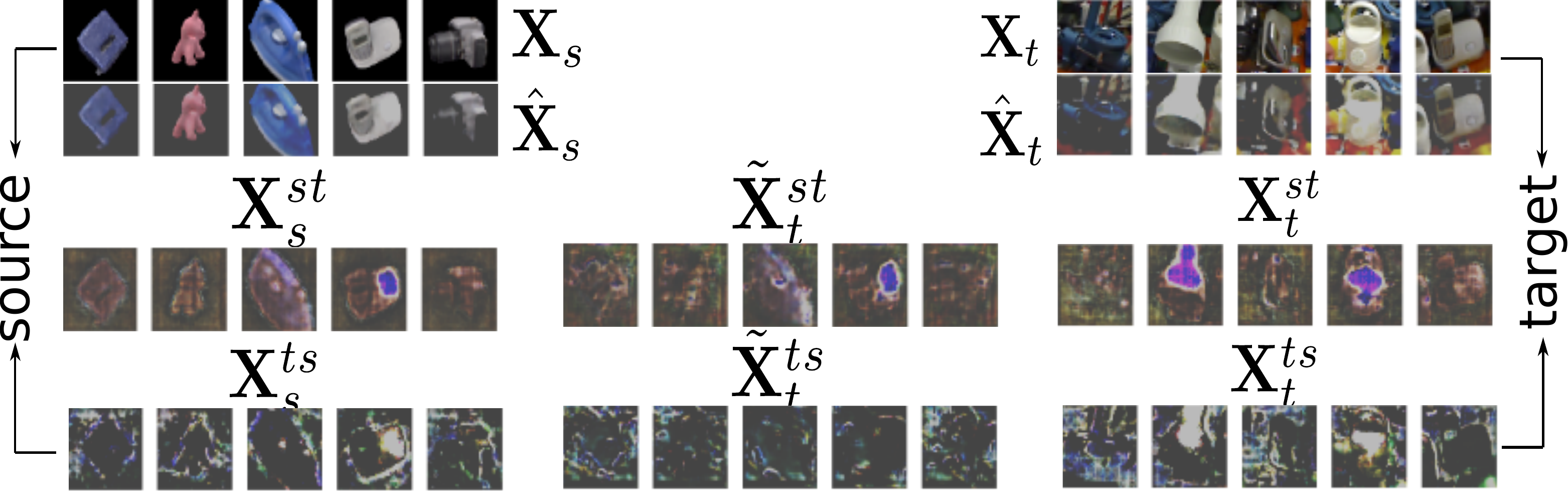}
        \caption{\small Linemod-CAD $\rightarrow$ Linemod-RW}
        \label{fig:linemod}
    \end{subfigure}
    \caption{\small Visualization of transition generations. For each scenario, the leftmost column is the source and its transition, and the rightmost is for the  target. During the experiments, the transitions of source are the `real player'. The adversarial generations for target transitions are in the middle column.}
    \label{fig:img_intermediates}   
\end{figure*}

\begin{figure*}[thbp]
    \centering
    \begin{subfigure}[t]{.32\textwidth}
        \centering
        \includegraphics[width=0.8\textwidth]{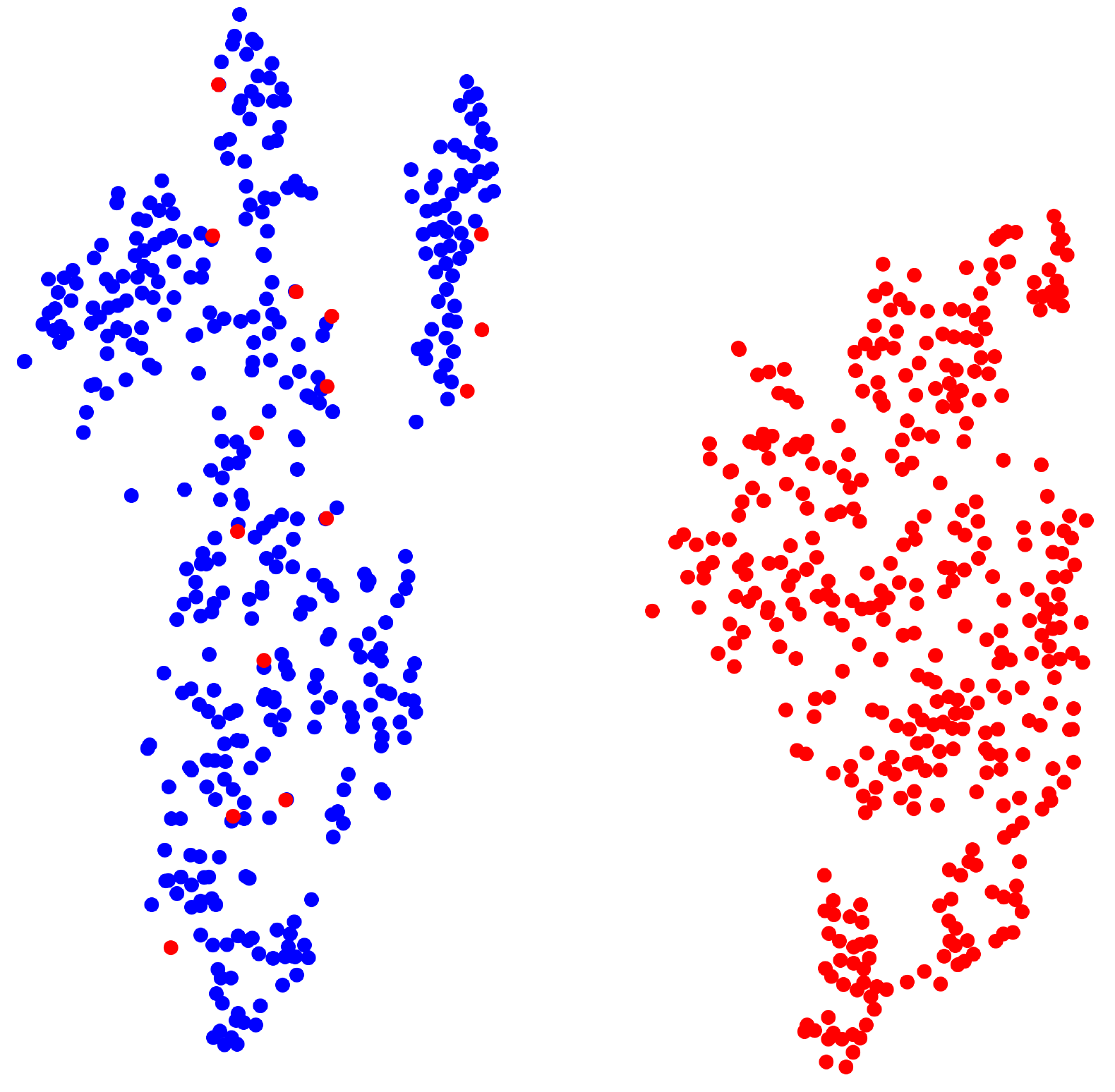}
        \caption{\small $\mathbf{X}_s$ (blue) vs. $\mathbf{X}_t$ (red)}
        \label{fig:tsne_mnist_mnistm_s_t}
    \end{subfigure}
    \begin{subfigure}[t]{.32\textwidth}
        \centering
        \includegraphics[width=0.8\textwidth]{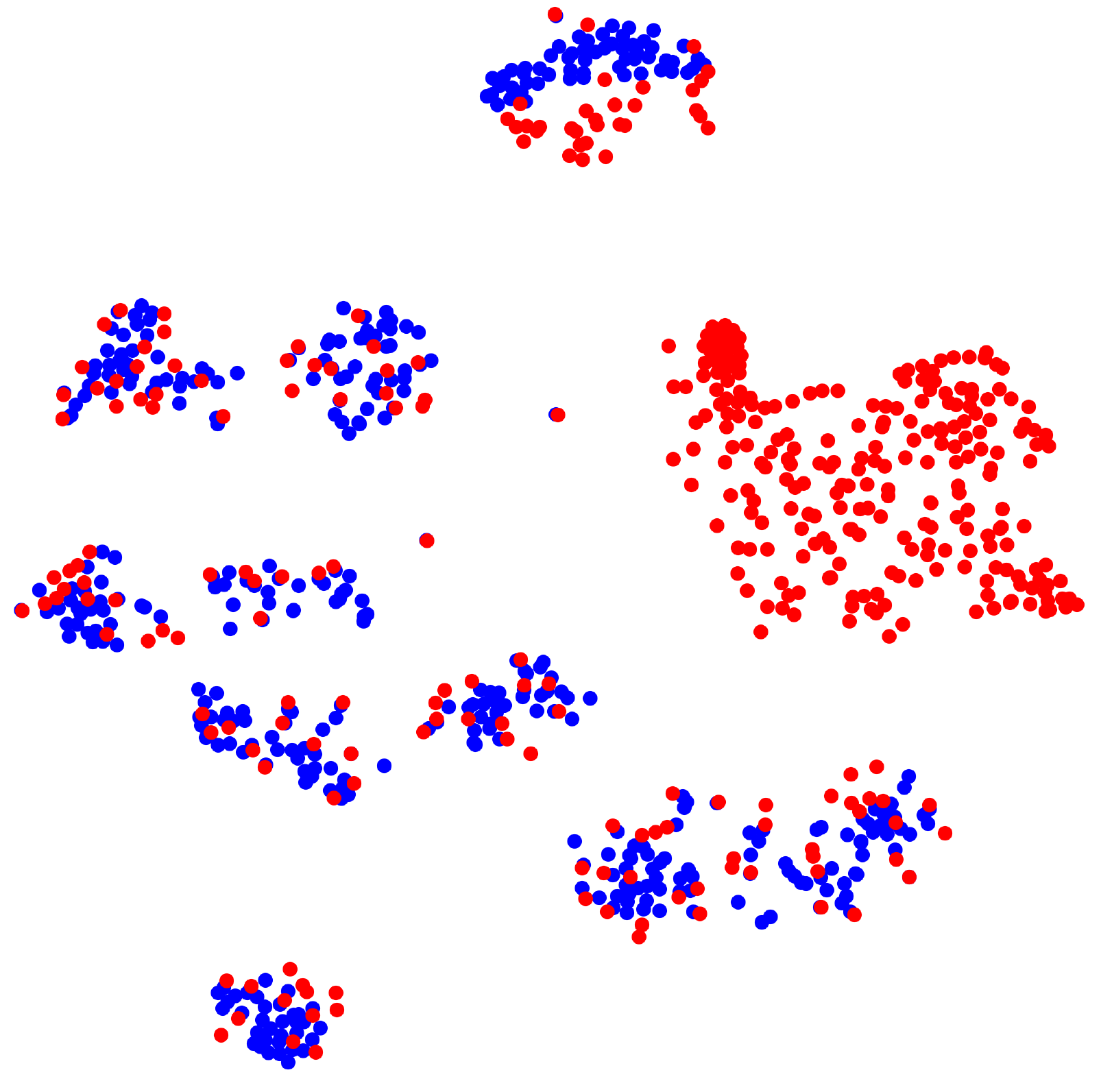}
        \caption{\small $\mathbf{X}_s^{st}$ (blue) vs. $\mathbf{X}_t^{st}$} (red)
        \label{fig:tsne_mnist_mnistm_st}
    \end{subfigure}
    \begin{subfigure}[t]{.32\textwidth}
        \centering
        \includegraphics[width=0.8\textwidth]{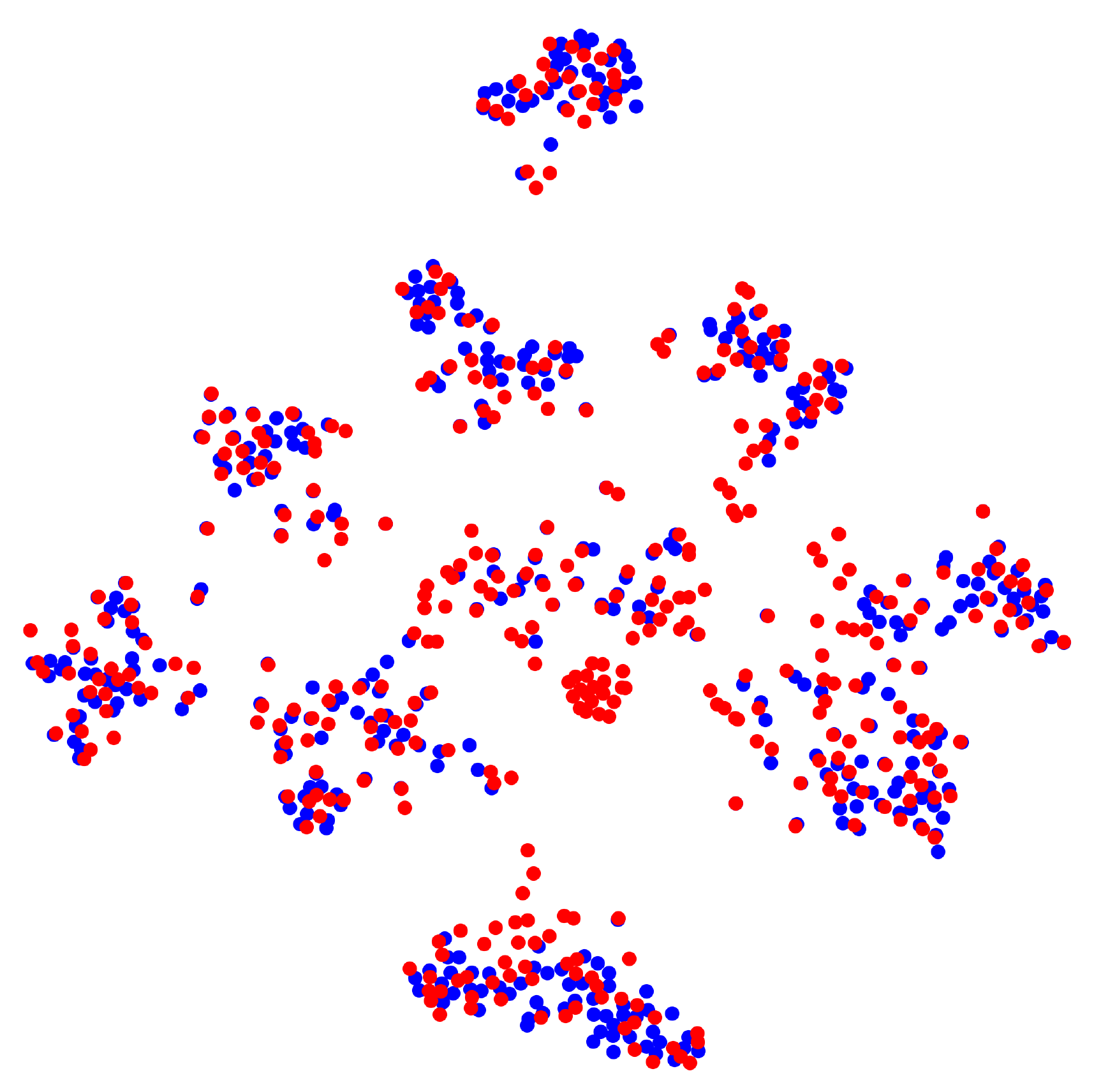}
        \caption{\small $\mathbf{X}_s^{st}$ (blue) vs. $\widetilde{\mathbf{X}}_t^{st}$} (red)
        \label{fig:tsne_mnist_mnistm_stg}
    \end{subfigure}
    \caption{\small  Visualization of domains and transitions using t-SNE for  MNIST$\rightarrow$MNIST-M. }
    \label{fig:tsne_mnist_mnistm}
\end{figure*}

\begin{figure*}[thbp]
    \centering
    \begin{subfigure}[t]{.3\textwidth}
        \centering
        \includegraphics[width=0.7\textwidth]{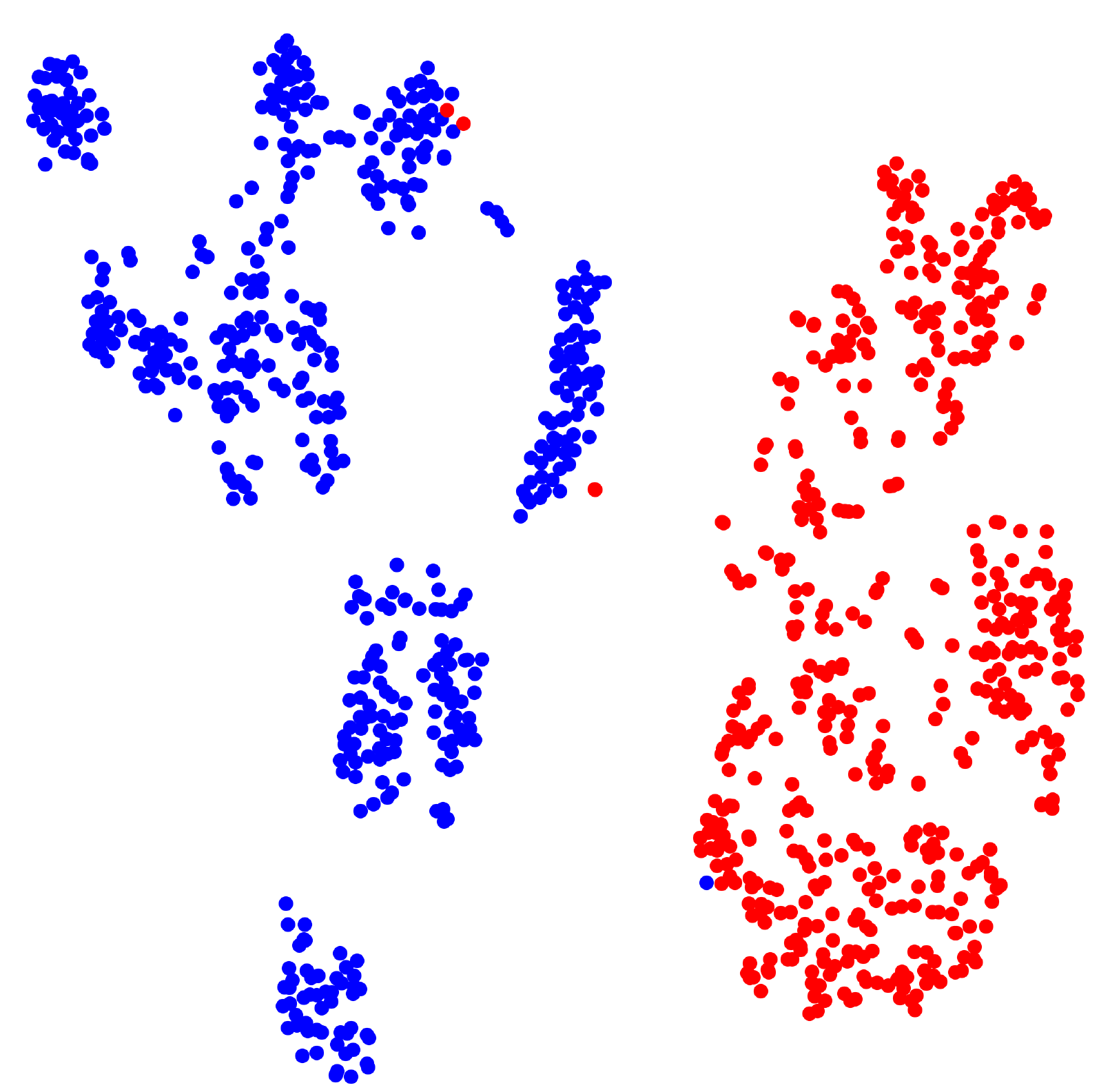}
        \caption{\small $\mathbf{X}_s$ (blue) vs. $\mathbf{X}_t$ (red)}
        \label{fig:tsne_mnist_usps_s_t}
    \end{subfigure}
    \begin{subfigure}[t]{.3\textwidth}
        \centering
        \includegraphics[width=0.7\textwidth]{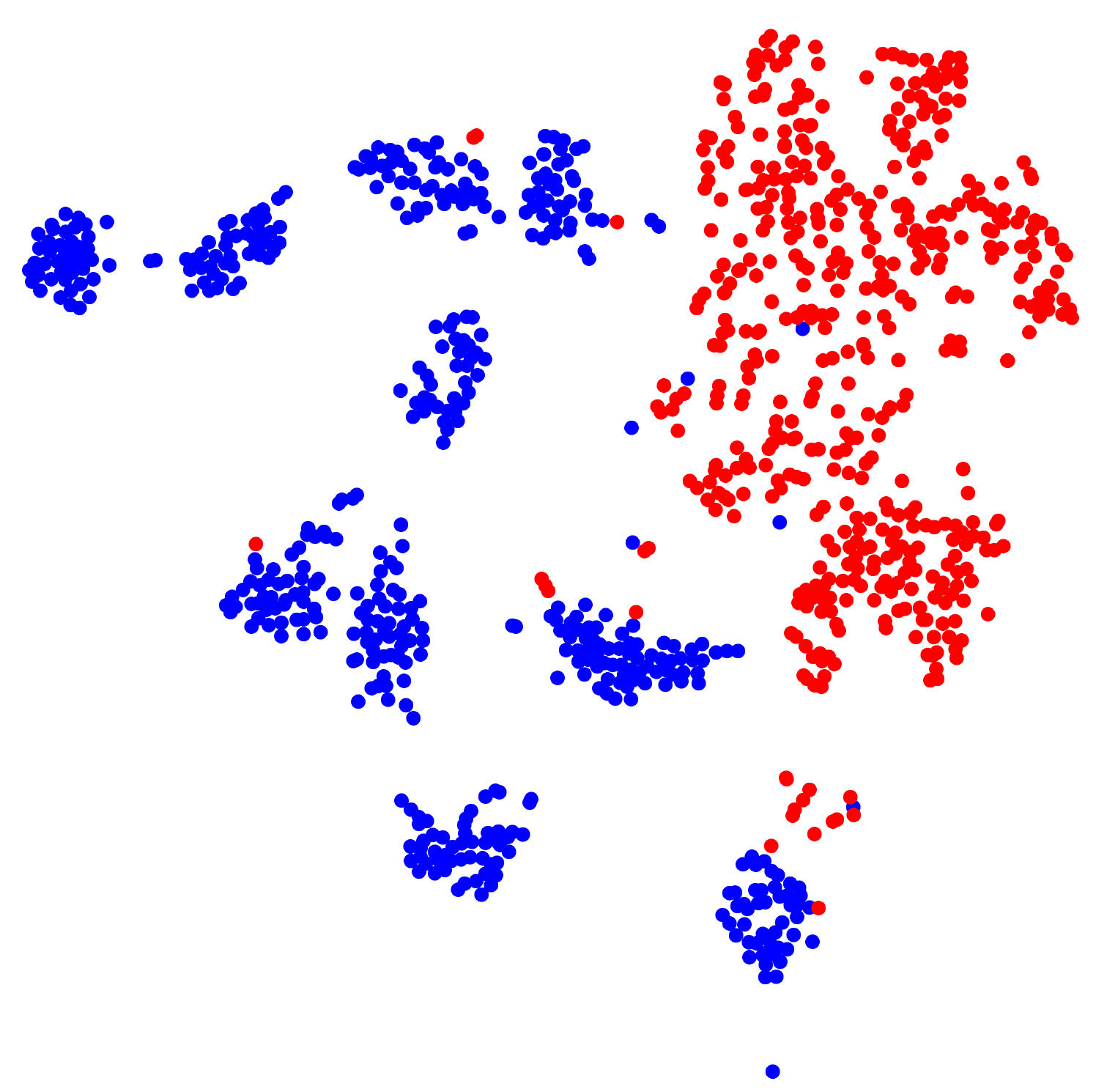}
        \caption{\small $\mathbf{X}_s^{st}$ (blue) vs. $\mathbf{X}_t^{st}$} (red)
        \label{fig:tsne_mnist_usps_st}
    \end{subfigure}
    \begin{subfigure}[t]{.3\textwidth}
        \centering
        \includegraphics[width=0.7\textwidth]{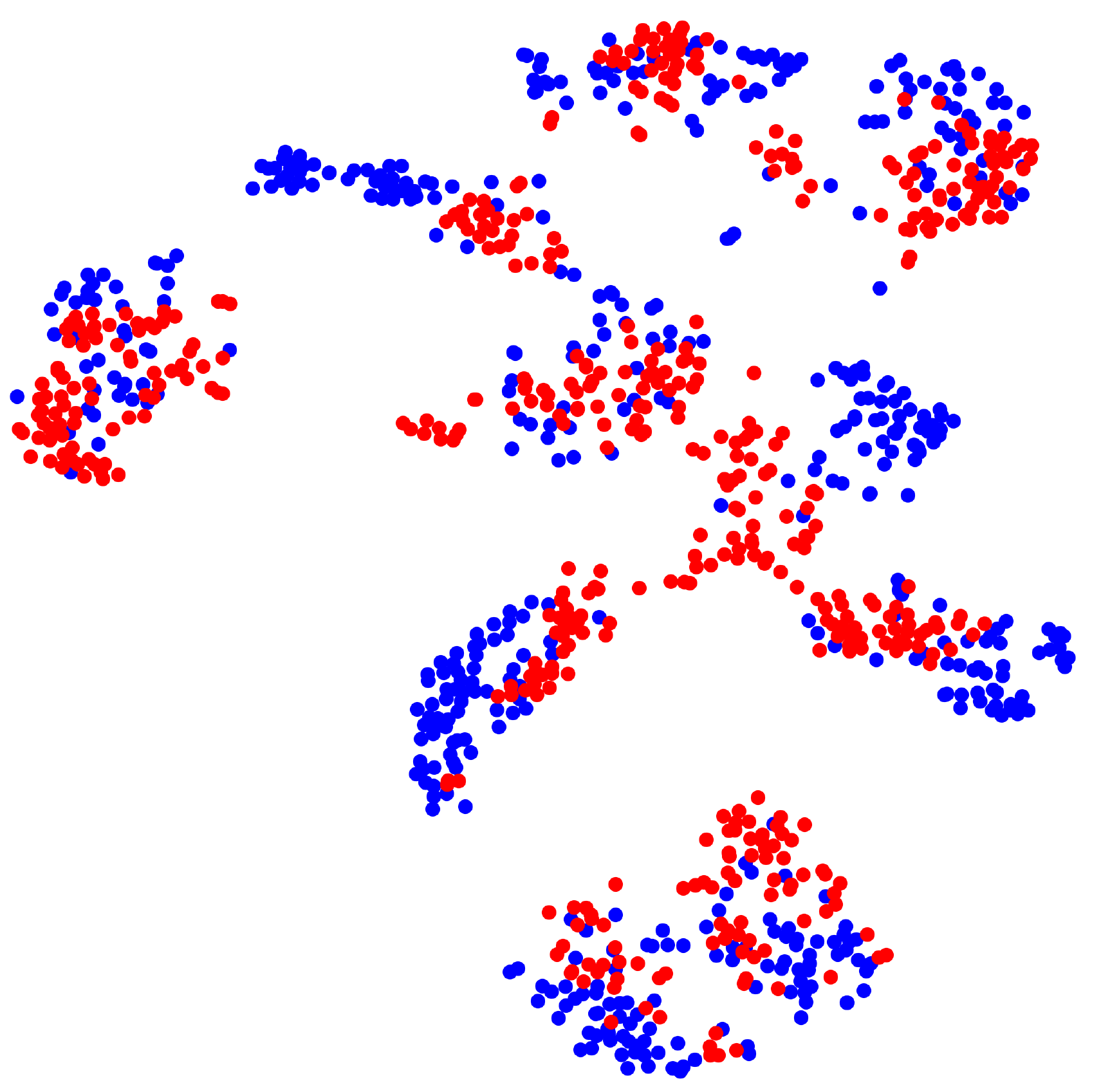}
        \caption{\small $\mathbf{X}_s^{st}$ (blue) vs. $\widetilde{\mathbf{X}}_t^{st}$} (red)
        \label{fig:tsne_mnist_usps_stg}
    \end{subfigure}
    \caption{\small  Visualization of domains and transitions using t-SNE for  MNIST$\rightarrow$USPS. }
    \label{fig:tsne_mnist_usps}
\end{figure*}

\begin{table*}[thbp]
    \centering
    \resizebox{1.0\textwidth}{!}{
        \begin{tabular}{|l|c|c|c|c|c|c|c|c|}
            \hline
            Source & MNIST & MNIST-M & MNIST & USPS & MNIST  & M-Digits & Fashion & Fashion-M \\
            Target & MNIST-M & MNIST & USPS  & MNIST & M-Digits & MNIST & Fashion-M & Fashion \\
            \hline\hline
            DATL w/o C ($\mathbf{X}^{st}$) &  0.821 & 0.935  & 0.946  & 0.938  & 0.895 & 0.902  & 0.835   & 0.865     \\
            DATL w/o C ($\mathbf{X}^{ts}$) & 0.923  &  0.840   & 0.902  &  0.930  & 0.853  & 0.851  & 0.803   & 0.832     \\
            DATL w/o GAN ($\mathbf{X}^{st}$) & 0.726  &  0.823   &  0.673  & 0.679  &  0.796  & 0.806  &  0.761  &  0.773  \\
            DATL w/o GAN ($\mathbf{X}^{ts}$) & 0.703  &  0.815   & 0.665   &  0.681  &  0.773  &  0.782  &  0.751  & 0.762  \\
            \hline      \hline
            DATL ($\mathbf{X}^{st}$) &  0.890 & \textbf{0.983}  & \textbf{0.961}  & 0.956  & \textbf{0.916} & \textbf{0.923}  & 0.853 & \textbf{0.917}     \\
            DATL ($\mathbf{X}^{ts}$) & \textbf{0.983}  &  0.871   & 0.943  &  0.953  & 0.883  & 0.892  & \textbf{0.886} & 0.903     \\
            \hline
    \end{tabular}}
    \caption{\small Ablation evaluation results for content similarity and GAN-based alignment in comparison with the full DATL framework.}
    \label{table:ablation_study}
\end{table*}

Since our model adopts a generative approach, we can have direct visual evaluation of the generated transitions. Some example  transitions are shown in Figure~\ref{fig:img_intermediates}, obtained after 100k mini-batch steps for the Fashion task and 50k for other three tasks.  In DATL, CGGS can generate the transitions with very similar appearance compared with the source and target domains, and the GANs are employed to move the transition and the generative transition pairs even closer. During transition generation, our model eliminates the strong noise of MNIST-M and Fashion-M. Though there are more complex textures in the Fashion task, the transitions of the Fashion scenario seem reasonable despite some information loss, possibly due to the complex textures and strong noise. The MNIST$\rightarrow$M-Digits scenario maintains the original content style, while the transitions display some style variation in the LineMod cases.

We adopt the t-SNE~\cite{Maaten2008} algorithm to further visualize the distribution of the transitions. The t-SNE visualization outcome for the MNIST$\rightarrow$MNISTM and MNIST$\rightarrow$USPS tasks are given in Figures \ref{fig:tsne_mnist_mnistm} and \ref{fig:tsne_mnist_usps}. In both figures we visualize three pairs of data or features: (a) the original data from the source and target domains, i.e., $\mathbf{X}_s$ and $\mathbf{X}_t$; (b) the transitions generated by VAEs with CGGS, i.e., $\mathbf{X}_s^{st}$ and $\mathbf{X}_t^{st}$; and (c) the transitions with adversarial alignment, i.e., $\mathbf{X}_s^{st}$ and $\tilde{\mathbf{X}}_t^{st}$. From the figures we can see that the original source and target data are well separated due to the domain gap despite some minor overlap (see Fig.~\ref{fig:tsne_mnist_mnistm_s_t}). For the CGGS-only transitions, although there is some reduction to the gap between the two, the alignment is lacking (especially in Fig.~\ref{fig:tsne_mnist_usps_s_t}). Finally,  Fig.~\ref{fig:tsne_mnist_mnistm_stg},~\ref{fig:tsne_mnist_usps_stg} are the t-SNE visualization of the transition $\mathbf{X}_s^{st}$ and its target-initiated counterpart $\tilde{\mathbf{X}}_t^{st}$ with adversarial learning, which give the best domain alignment. Overall, these visualization outcomes clearly justify the DATL framework design using both CGGS and adversarial alignment. 

\subsubsection{Ablation Study}
\label{subsubsec:ablation_study}

Ablation studies are conducted for the content similarity loss and the GAN-based alignment. 

\noindent
\textbf{Content similarity.} To evaluate the potential effect of employing the content similarity loss in our model, we conduct the adaptation tasks without applying content similarity, with the ablated model denoted by ``DATL w/o C''. From Table~\ref{table:ablation_study}, we can see that the full model with content similarity outperforms its ablated counterpart. This confirms the necessity of incorporating content similarity.

\noindent
\textbf{GAN-based alignment.} To evaluate the effect of the GAN-based domain alignment in our model, we test the adaptation performance of an ablated model using the transitions generated by CGGS (without further adversarial learning). A classifier is trained by the $\mathbf{X}_s^{st}$ data, and then is tested on the $\mathbf{X}_t^{st}$ data. Table~\ref{table:ablation_study} gives the results of the evaluations. We can see that the performance of the ablated model is better than the source-only scenario, but much worse than that of the full DATL framework. 

\subsubsection{Model Analysis}
\label{subsubsec:model_analysis}

\begin{table*}[thbp]
    \centering
    \resizebox{0.9\textwidth}{!}{
        \begin{tabular}{|l|c|c|c|c|}
            \hline
            \multirow{2}{*}{Home Task}& \multicolumn{4}{c|}{Migrated Tasks} \\ \cline{2-5}
             &  MNIST$\rightarrow$MNIST-M & MNIST$\rightarrow$USPS & MNIST$\rightarrow$M-Digits & Fashion$\rightarrow$Fashion-M\\
            \hline\hline
            MNIST$\rightarrow$MNIST-M  &\textbf{0.890(0.983)} & 0.958(0.945)  & 0.915(0.853)  &  0.762(0.730) \\
            MNIST$\rightarrow$USPS &0.915(0.859)  & \textbf{0.961(0.943)} & 0.882(0.914)  & 0.605(0.587)   \\
            MNIST$\rightarrow$M-Digits &0.843(0.928) & 0.944(0.958) & \textbf{0.916(0.883)} & 0.613(0.593)  \\
            Fashion$\rightarrow$Fashion-M & 0.925(0.881) &0.932(0.935) & 0.825(0.913)  & \textbf{0.853(0.886)}  \\
            \hline
        \end{tabular}
    }
    \caption{\small Performance of cross-task generalization using the $\mathbf{X}^{st}$ channel (results of using $\mathbf{X}^{ts}$  shown in  parentheses).}
    \label{table:generalization}
\end{table*}

Some further experiments are done to evaluate our model.

\noindent
\textbf{$L2$ distance evaluation of transitions:} In addition to the visualization results given in Figures 6 and 7, we evaluate the $L2$ distance between the original source and target ($\mathbf{X}_s, \mathbf{X}_t$), the transitions before adversarial training ($\mathbf{X}_s^{st}, \mathbf{X}_t^{st}$), also the transitions after adversarial training($\mathbf{X}_s^{st}, \widetilde{\mathbf{X}}_t^{st}$).  Specifically, we calculate the average MSE over the test batches. From the results in Table~\ref{table:l2-distance}, we can see that the distance between the $\mathbf{X}_s^{st}$ and $\widetilde{\mathbf{X}}_t^{st}$ is smallest; and the distance between the original source and target domains is the largest. The distance of the transitions without adversarial training is moderate. These $L2$ distance results are consistent with the visualization outcome, clearly demonstrating the initial reduction of gaps between the CGGS-generated transitions, and the further reduction obtained through transitions' adversarial alignment. 

\begin{table}[htbp]
    \centering
    \resizebox{0.3\textwidth}{!}{
    \begin{tabular}{|c|c|c|}
        \hline
        Source & MNIST &  MNIST \\
        Target & MNISTM &  USPS \\
        \hline\hline
        $L2(\mathbf{X}_s, \mathbf{X}_t)$ & 0.235 & 0.258  \\
        $L2(\mathbf{X}_s^{st}, \mathbf{X}_t^{st})$ & 0.075 &  0.085 \\
        $L2(\mathbf{X}_s^{st}, \widetilde{\mathbf{X}}_t^{st})$ & 0.016 & 0.034 \\
        \hline
    \end{tabular}}
    \caption{Average $L2$ distances of domain data and related transitions.}
    \label{table:l2-distance}
\end{table}

\begin{figure}[thbp]
    \begin{subfigure}[t]{0.48\textwidth}
        \centering
        \includegraphics[width=0.9\textwidth]{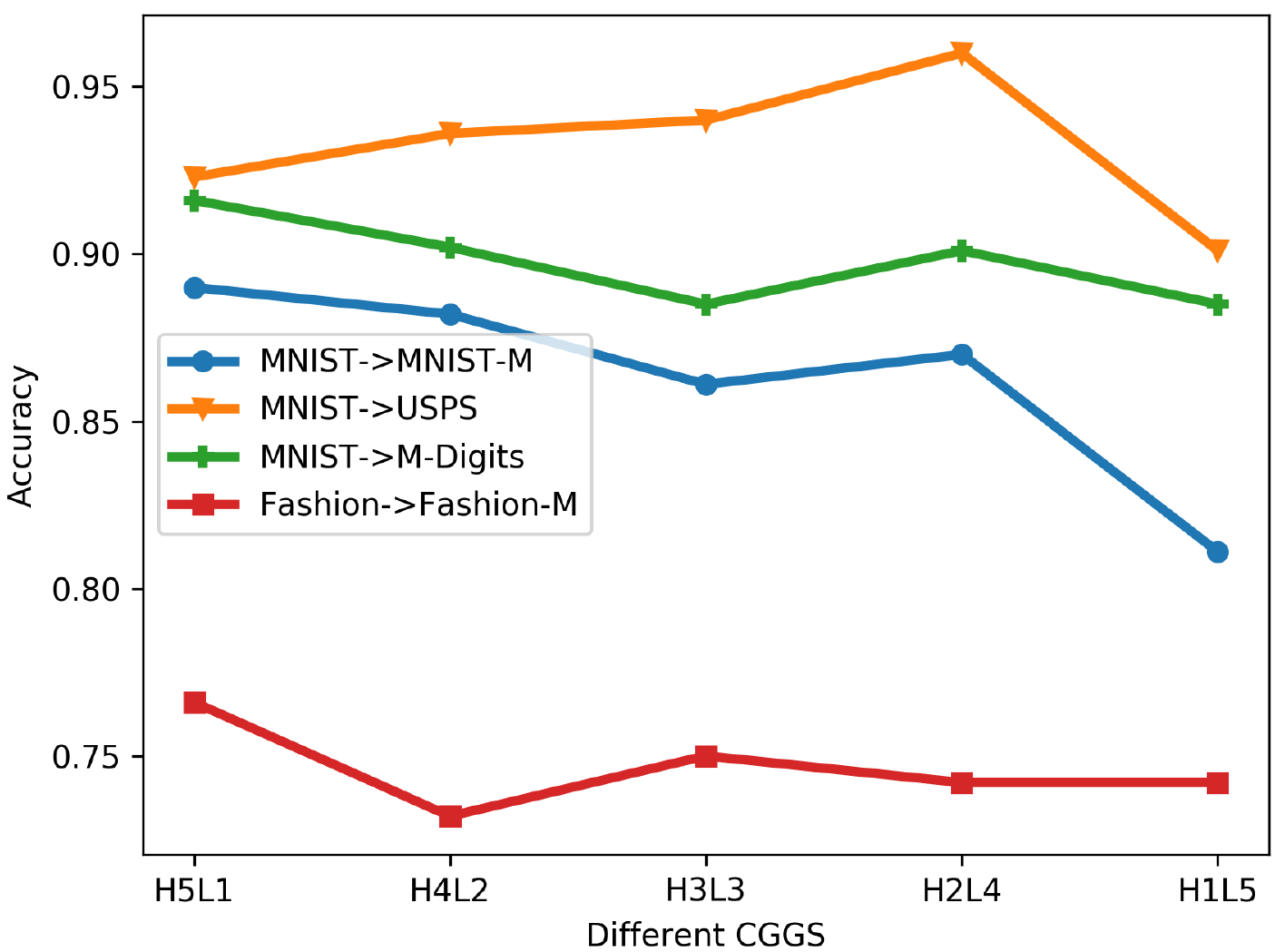}
        \caption{\small $\mathbf{X}^{st}$ channel}
        \label{fig:cgrs_st}
    \end{subfigure}
    \begin{subfigure}[t]{.48\textwidth}
        \centering
        \includegraphics[width=0.9\textwidth]{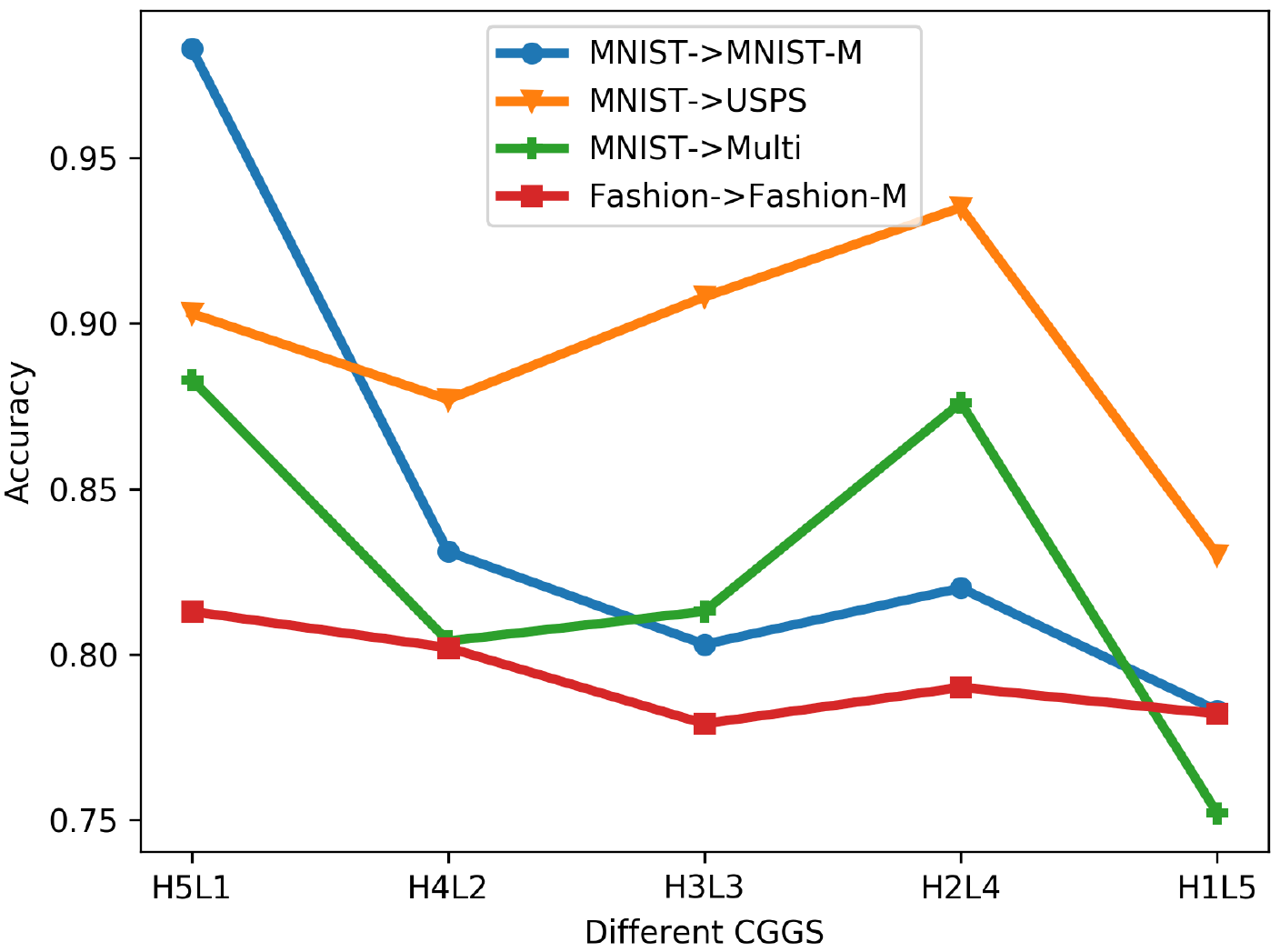}
        \caption{\small $\mathbf{X}^{ts}$ channel}
        \label{fig:cgrs_ts}
    \end{subfigure}
    \caption{\small Adaptation performance of different CGGS settings.}
    \label{fig:cgrs}
\end{figure}

\noindent
\textbf{Sensitivity of CGGS settings:} As CGGS plays an important role in the proposed model,  we evaluate the effect of varying the structure of CGGS. During the experiments, we use a fix depth of network (6 layers) for the generation process. We apply various settings for splitting the high-level and low-level decoder stacks for CGGS. For example, H5L1 denotes a scheme of using 5 layers for the high-level stack and 1 layer for the low-level. The results of changing the CGGS setup for four tasks are shown in Figure~\ref{fig:cgrs}. It can be seen that for the channel $\mathbf{X}^{st}$ in MNIST$\rightarrow$MNIST-M and Fashion$\rightarrow$Fashion-M tasks, the highest accuracies are reached at H5L1, and for MNIST$\rightarrow$USPS and MNIST$\rightarrow$M-Digits tasks, there is a peak value at H2L4; the performance variation, however, seems moderate. The $\mathbf{X}^{ts}$ channel somehow seems relatively more sensitive to the change of CGGS settings.  

\noindent  
\textbf{Cross-task generalization:} So far we have evaluated DATL's performance \textit{within} individual domain adaptation tasks. To further explore the generalization ability of DATL, we have also experimented with cross-task generalization, i.e. using the DATL models pre-trained in one task (`home task') for another adaptation task (`migrated task'). In our experiments, $4\times 4$ pair-wise task migration scenarios are evaluated. These DATL models are trained with a trade-off H4L2 grafted structure according to the sensitivity analysis. During the experiments, we keep the CGGS fixed, then fine-tune the adversarial alignment within the new tasks. The results are shown in  Table~\ref{table:generalization}. Although in general there is a slight reduction to the home-task adaptation accuracy (shown as the diagonal entries in bold in the table), the performance of adaptation to the migrated tasks are comparable. Specifically,  MNIST$\rightarrow$MNIST-M and Fashion$\rightarrow$Fashion-M as home tasks adapt well to other tasks, while  MNIST$\rightarrow$USPS and MNIST$\rightarrow$M-Digits get a lower accuracy from the  Fashion$\rightarrow$Fashion-M task. These results seem not surprising, as the learning outcome from a difficult task (such as Fashion$\rightarrow$Fashion-M) is more likely to generalize well to easier tasks.

\subsubsection{Semi-supervised Learning}
\label{subsubsec:semi_supervised_evaluation}

Finally, we evaluate the performance of our model for semi-supervised learning. Under this scenario, it is assumed that we can get a small number of labeled target samples. Following the treatment in~\cite{Bousmalis2017}, we choose 1000 or 2000 samples from every category in the target domain with labels. These are augmented to the source domain data and used for training. The results are shown in Table~\ref{table:semi-supervised}. The adaptation performance is better than the unsupervised scenario, whereas having 2000 target samples for data augmentation will further improve the performance. 

\begin{table}[!t]
    \centering
    \resizebox{0.45\textwidth}{!}{
        \begin{tabular}{|l|c|c|c|c|}
            \hline
            Source & MNIST &  MNIST & MNIST & Fashion\\
            Extra & MNIST-M & USPS & M-Digits & Fashion-M\\
            \hline\hline
            1000 & 0.988  & 0.966 & 0.925 & 0.865 \\
            2000 & 0.990 & 0.970 & 0.932  & 0.892 \\
            \hline
    \end{tabular}}
    \caption{\small  Classification performance of DATL with semi-supervised learning  for four tasks.}
    \label{table:semi-supervised}
\end{table}


\section{Conclusion} \label{sec:conslusion}


In this paper, we have proposed a novel unsupervised domain adaptation model based on cross-domain transition generation and label alignment using adversarial networks. In particular, cross-grafted generative stacks from VAEs of different domains are constructed to generate bidirectional transitions, which are further aligned using generative adversarial learning. The domain adaptation task hence is transformed to constructing an effective mapping of the cross-domain transitions onto the label space of the original source domain, a methodology we believe contributes to its robust performance in domain adaptation tasks. This is verified by the extensive empirical results we have obtained from a number of benchmark tasks and supported by ablation studies as well as the visualization of the transition spaces. 

Our experiment results reveal that the proposed DATL method can maintain stable performance when we vary the settings for stack splitting and crafting. There seems to be a tendency to favor a higher ratio of high-level to low-level layers when the domains contain similar contents but different background, while adaptation tasks with similar background but different content favor more low-level layers. The flexibility on the CGGS setting may be a good feature for performance tuning when dealing with real-world applications. 

Another interesting observation is that DATL has a very good cross-task generalization ability. The model trained by one task can be employed for domain adaptation in another task. This demonstrates a merit of our method for practical applications. We believe it also raises a challenging, new perspective for domain adaption tasks that deserves more future work. 

Finally, while both transition channels in DATL are well aligned and usable for domain adaption tasks, from our experiments it seems the $\mathbf{X}^{st}$ channel displays better classification performance more often. This may be coincidental as our DATL framework is symmetric, and our domain adaptation tasks also report the most symmetric performance. As another direction for future work, it may be possible to design an ensemble method that combines classification outcomes obtained under different CGGS settings using both transitional channels.


\ifCLASSOPTIONcaptionsoff
  \newpage
\fi

{\small
	\bibliographystyle{IEEEtran}
	\bibliography{cgrs}}

\end{document}